\documentclass[letterpaper, 10pt, conference]{ieeeconf}
\IEEEoverridecommandlockouts
\usepackage{csquotes} 
\usepackage{relsize} 
\usepackage{amssymb} 
\usepackage[nice]{nicefrac}
\usepackage[free-standing-units]{siunitx} 
\usepackage{soul} 
\usepackage{colortbl} 
\usepackage{graphicx,color}
\usepackage{enumerate}
\definecolor{gray}{rgb}{0.7,0.7,0.7}
\definecolor{orange}{RGB}{255,165,0}
\newcommand{\graycell}{\cellcolor{gray}}

\newcommand{\rc}{{RoboCup@Work}}

\newcommand{\auxnet}{\textsc{Auxnet}}
\newcommand{\approach}{\textsc{Converge-Fast-\auxnet}}

\title{Fast Convergence for Object Detection\\by Learning how to Combine Error Functions}

\author{\authorblockN{Benjamin Schnieders}
\authorblockA{Department of Computer Science\\
University of Liverpool\\
L69 3BX Liverpool, United Kingdom\\
Email: bsc@liv.ac.uk}
\and
\authorblockN{Karl Tuyls}
\authorblockA{Department of Computer Science\\
University of Liverpool\\
L69 3BX Liverpool, United Kingdom\\
Email: ktuyls@liv.ac.uk}
}

\begin{document}
\clearpage
\vspace*{\fill}
\begin{minipage}{\textwidth}
\begin{center}
\begin{minipage}{.6\textwidth}
\huge
\centering {\copyright} 2018 IEEE.\\
\large
\vspace{0.5cm}
Personal use of this material is permitted. Permission
from IEEE must be obtained for all other uses, in any current or future
media, including reprinting/republishing this material for advertising or
promotional purposes, creating new collective works, for resale or
redistribution to servers or lists, or reuse of any copyrighted
component of this work in other works.\\
\vspace{1.0cm}
Accepted version.
\end{minipage}
\end{center}
\end{minipage}
\vfill
\clearpage

\maketitle
\thispagestyle{empty}
\pagestyle{empty}
\begin{abstract}
In this paper, we introduce an innovative method to improve the convergence speed and accuracy of object detection neural networks. Our approach, {\approach}, is based on employing multiple, dependent loss metrics and weighting them optimally using an on-line trained auxiliary network. Experiments are performed in the well-known {\rc} challenge environment.  A fully convolutional segmentation network is trained on detecting objects' pickup points. We empirically obtain an approximate measure for the rate of success of a robotic pickup operation based on the accuracy of the object detection network. Our experiments show that adding an optimally weighted Euclidean distance loss to a network trained on the commonly used Intersection over Union (IoU) metric reduces the convergence time by 42.48\%. The estimated pickup rate is improved by 39.90\%. Compared to state-of-the-art task weighting methods, the improvement is 24.5\% in convergence, and 15.8\% on the estimated pickup rate.
\end{abstract}

\setcounter{footnote}{0}
\section{Introduction}
Although Deep Convolutional Neural Networks (DCNNs) form the state of the art in object detection~\cite{schmidhuber2015deep}, they typically require a high number of training iterations to converge.
In object detection and robotic pick-up frameworks like the \emph{Factory of the Future}~\cite{lasi2014industry}, a combination of low training times and accurate object detection is vital to ensure minimal turnaround times.

\begin{figure}[tb]
	\centering
	\includegraphics[width=1\linewidth]{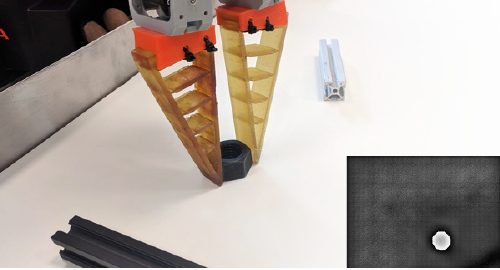}
	\label{fig:pickup}	
	\caption{Picking an object from the position indicated by the activations of the DCNN (lower right). The dominant axis of the activations is regarded as the facing of the object, and the gripper oriented perpendicular to it.}
\end{figure}

This paper introduces an innovative method for fast convergence of DCNNs performing object detection tasks, using an auxiliary network dubbed {\approach}. Our method combines multiple, synergistic loss functions to improve both the accuracy and convergence speed of DCNNs.
We use an image segmentation DCNN for object detection, and use the auxiliary neural network to learn optimal weights for its multiple loss functions during training. No pre-processing is required on the error metrics, and no additional hyper-parameters are introduced.
The approach is tested in a {\rc}~\cite{kraetzschmar2014robocup} setup, simulating the requirements for the Factory of the Future. Figure~\ref{fig:pickup} shows a typical pickup operation in the RoboCup@Work competition.
Results illustrate that {\approach} outperforms the state of the art in accuracy, convergence speed, and stability.
Necessary training iterations are reduced on average by $24.5\percent$, or roughly an hour of training on a NVIDIA Titan X. The probability of a successful picking operation is improved by $15.8\percent$ on average ($9\percent$ absolute difference). The IoU measure is improved by $10.5\percent$.
Our main contributions can thus be summarized as follows:
\begin{enumerate}[i]
\item the introduction of the {\auxnet} approach, which learns optimal weights on-line in a multi-objective environment,
\item the combination of the commonly used Intersection over Union (IoU) metric and a Euclidean distance loss for faster convergence,
\item providing a labeled object classification and detection benchmark dataset, which is publicly available.\footnote{See \tt{https://airesearch.de/ObjectDetection@Work/}}
\end{enumerate}

The remainder of this paper is organized as follows: Section~\ref{sec:bg} provides background in object detection and presents related work.
Section~\ref{sec:impl} introduces the {\approach} approach and describes the introduced loss functions in detail.
Section~\ref{sec:exp} describes the experimental setup.
Finally, Section~\ref{sec:res} presents and compares the experimental results and Section~\ref{sec:conc} concludes the paper.

\section{Background and Related Work}
\label{sec:bg}
Object detection is a well-researched task in the field of computer science and artificial intelligence. The purpose is to output a list of pre-trained objects that are present in a given image, along with their positions in the image~\cite{guo2016deep}.
Major breakthroughs using DCNNs in the last years have advanced the performance of neural network object predictors such that DCNNs are now the leading type of classifier in many image recognition~\cite{deng2009imagenet} or detection benchmarks, most notably, the Pascal VOC Challenge~\cite{everingham2010pascal}.
State-of-the-art approaches such as Faster-RCNN~\cite{ren2017faster} or YOLO9000~\cite{redmon2017yolo9000} pass an image through the network once, then propose regions for bounding boxes which are subsequently refined.
Another way to detect objects, avoiding bounding box creation, is using semantic segmentation~\cite{long2015fully}. A semantic segmentation network features input and output layers of the same dimension, with each output pixel denoting which of the trained classes it belongs to. This effectively allows a different object to be predicted at every pixel. Further processing, such as clustering, is required to obtain the position of each object. The DCNN used in this research is based on the segmentation approach.

Multi-task learning can be used to improve both training time and prediction accuracy by learning a shared representation for multiple objectives simultaneously~\cite{caruana1998multitask}.
Multiple tasks can be learned by a single network by combining the respective task loss functions.
Commonly, multi-task learning is employed to learn related, but independent tasks by a shared model, such as object detection and classification~\cite{sermanet2014overfeat}.
Combining error functions typically consists of scaling, weighting and finally summing them. These weights are conventionally modeled as hyperparameters~\cite{girshick2015fast}.
Kendall et al.~\cite{kendall2017multi} present a statistically sound way to learn optimal weights, assuming that the epistemic error in the model will be eliminated with enough learning, and the dominant form or error remaining will be the homoscedastic uncertainty, which can be captured by measuring the variance of the loss over time. An error function with lower variance will then gain a higher relative weight, and vice versa. We will refer to this weighting method as \emph{KGC-weighting}, after the initials of the authors. The combined loss $\mathcal{L}_{KGC}$ is defined in terms of the individual losses $\mathcal{L}_i$, and their respective variances $\sigma^2_i$ are defined as shown in Equation~\ref{eq:kgc}.

\begin{equation}
\label{eq:kgc}
\mathcal{L}_{KGC} = \sum_{i}\frac{1}{2\sigma^2_i}\mathcal{L}_i+\log\sigma^2_i
\end{equation}
An immediate limitation of this approach is that functions with extremely low variances, such as an IoU with little or no overlap, may not train well as they can produce exploding gradients. Section~\ref{sec:res:owl} presents two approaches to address this issue: $KGC_{+\epsilon}$ and $KGC_{\nicefrac{}{Mean}}$-weighting.

\section{Approach}
\label{sec:impl}
In this work, we will show that combining mutually \emph{dependent} error functions can provide a significant improvement in both convergence time and the resulting classifier accuracy. Common practice is to train multi-objective networks using mutually \emph{independent} error functions, that is, when training one independent function, the values of the others are not affected.

While~\cite{caruana1998multitask} hints that tasks can be too similar to gain improved performance from training a shared model, we will demonstrate that in an object detection framework, training a shared model minimizing two mutually dependent loss functions, i.e., loss functions that are dependent in such a way that training one may also reduce the other, provides a significant improvement over learning just one.

\begin{figure}[tb]
	\centering
	\includegraphics[width=0.8\linewidth]{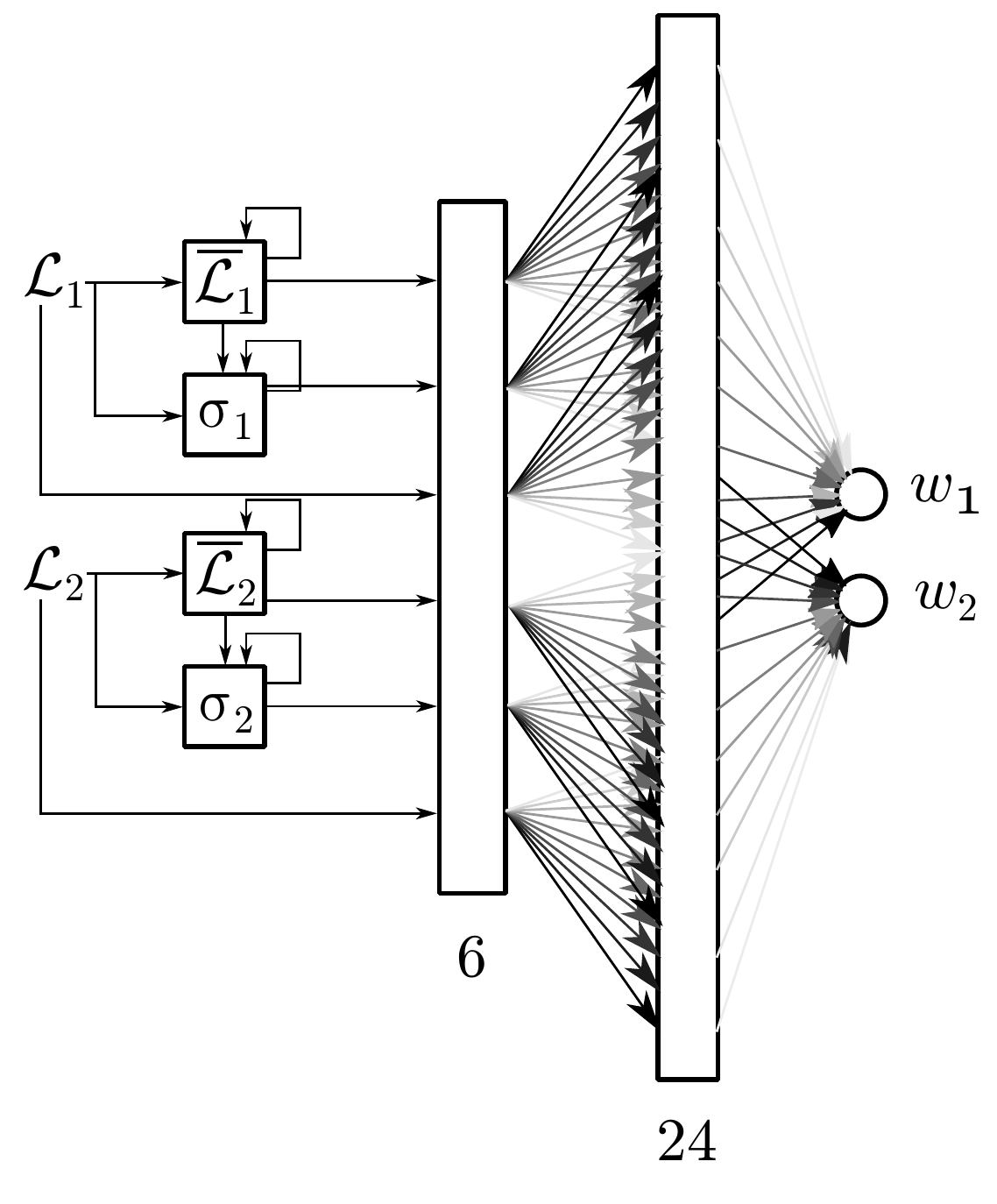}
	\caption{The layout of {\auxnet}. The current losses $\mathcal{L}_i$ are averaged to $\overline{\mathcal{L}}_i$ over the course of the training session using exponential moving averages. The standard deviation $\sigma_i$ is obtained likewise. The fully connected hidden layer produces output weights, which are inserted into Equation~\ref{eq:auxnet}. The network optimized to the gradient of the DCNN error reduction, as detailed in Equation~\ref{eq:auxnet_err}.}
	\label{fig:auxnet}
\end{figure}

We then present {\approach}, a method that models the learning of optimal error function weights as an auxiliary task~\cite{zhang2014supervised}. The joint error function can be described as minimizing the scaled, summed up error of all the available error functions, while optimizing their weights for fastest reduction. As an auxiliary network, we use a fully connected neural network with input features being the current value, the average, and standard deviation of every error function. The hidden layer consists of 24 ReLU neurons, which are combined into the two weights used to scale the two error functions. Figure~\ref{fig:auxnet} displays the auxiliary network layout. While the number of neurons in the hidden layer may be seen as an additional hyper-parameter, they should only depend on the number of error functions to be weighted, not the type or scale of the functions used.

The \emph{total loss} of the DCNN, $\mathcal{L}_{Total}$ is derived from summing all weighted individual functions. The learned weights for every individual error function are applied to the respective loss value in a manner similar to \emph{KGC-weighting}, i.e., dividing each loss value $\mathcal{L}_i$ by the respective weight $w_i$ and adding the natural logarithm of the weight, as shown in Equation~\ref{eq:auxnet}.
Weighting the terms as shown introduces self-regulating properties of the weights compared to a simple multiplication with the loss.
The auxiliary network is maximizing the rate of decline in \emph{total loss}, the term to be minimized is provided in Equation~\ref{eq:auxnet_err}.
$\mathcal{L}_{Total}$ and $\mathcal{L}_{\textsc{Auxnet}}$ are optimized with two different Adam instances, both initialized with the same parameters as listed in Section~\ref{sec:bg}.
The individual error functions to be weighted and subsequently collectively minimized are presented in Section~\ref{sec:errfunc}.

\begin{equation}
\label{eq:auxnet}
\mathcal{L}_{Total} = \sum_{i}\frac{1}{w_i}\mathcal{L}_i+\log w_i
\end{equation}

\begin{equation}
\label{eq:auxnet_err}
\mathcal{L}_{\textsc{Auxnet}} = \frac{\mathcal{L}_{Total} - \overline{\mathcal{L}_{Total}}}{\overline{\mathcal{L}_{Total}}}
\end{equation}

\subsection{Network}
\begin{figure}[tb]
	\centering
\includegraphics[width=1\linewidth]{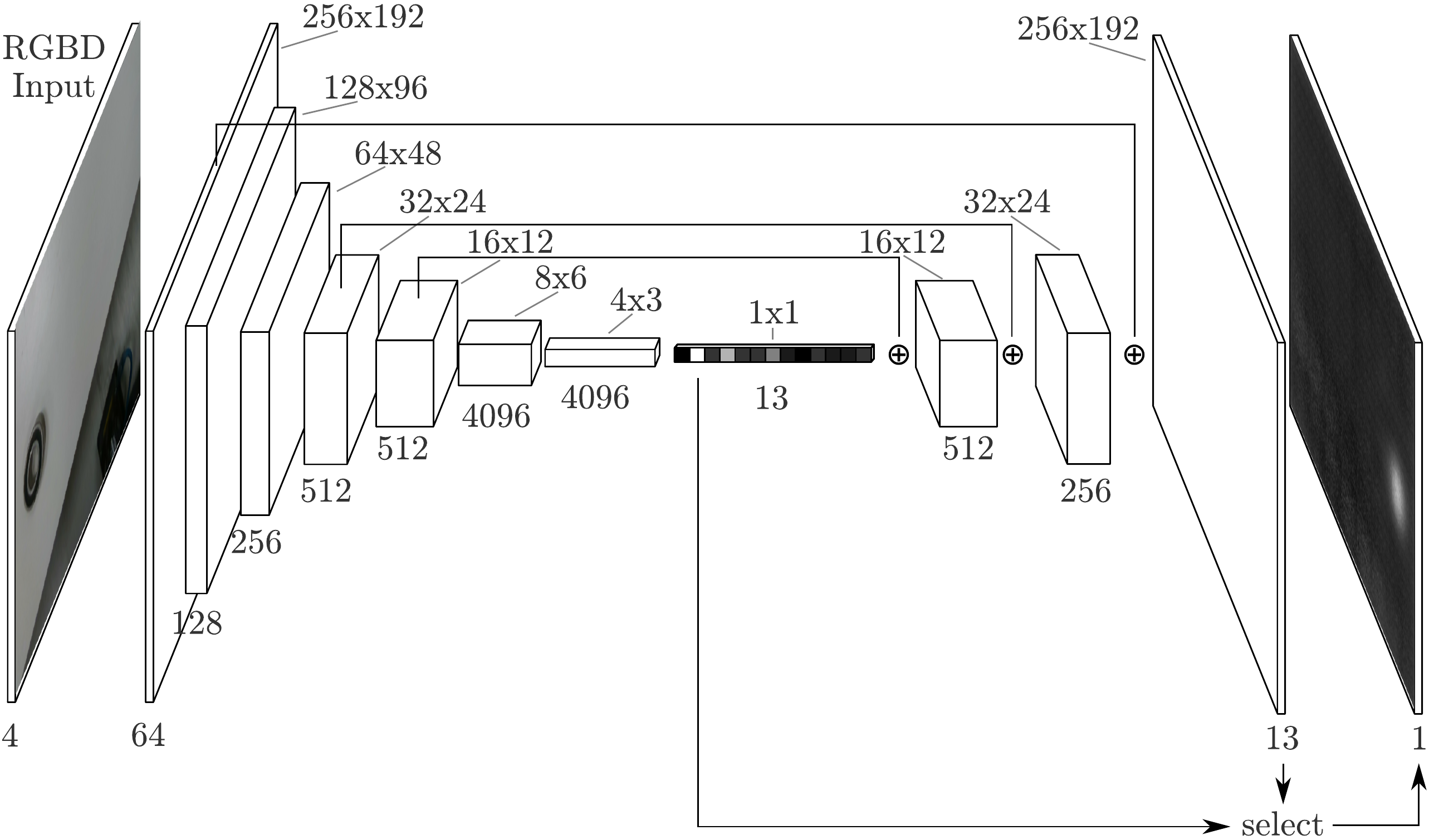}

	\caption{The network layout used. Numbers above the layers indicate pixel dimensions, numbers below the image denote the number of channels or filters. All layers are convolutional/inverse convolutional, using ReLU activation functions. It is trained on either minimizing the IoU or distance error for a queried object on the corresponding output layer, or on a combination of both using Equation~\ref{eq:auxnet}.}
	\label{fig:hourglass_net}

\end{figure}

The layout of the DCNN used was inspired by the networks described in~\cite{simonyan2014very}, but was extended to feature the traditional hour-glass shape of semantic segmentation networks~\cite{long2015fully}. The derived network features four instead of the common three input channels, with the fourth channel encoding distance information obtained by using an Intel SR300 3D camera for construction of the dataset. The number of filters per layer was tweaked by hand, and the number of output layers reflects the number of different objects to be distinguished by the network.
Before being presented to the network, all images are scaled down from the native resolution of $640 \times 480$ pixels to a more manageable size of $256 \times 192$ pixels.
All activation functions are ReLUs. Regularization is introduced in the form of dropout regularization, with a $15\percent$ dropout chance in the convolutional layers 6 and 7. The innermost layer doubles as a classification vector. If a specific object is to be retrieved, only the activations on the corresponding layer are taken into account.
A detailed description of the network can be found in Figure~\ref{fig:hourglass_net}.

\subsection{Error Functions}
\label{sec:errfunc}
\begin{figure}[tb]
	\centering
	\includegraphics[width=0.49\linewidth]{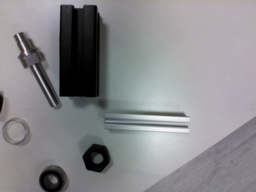} \includegraphics[width=0.49\linewidth]{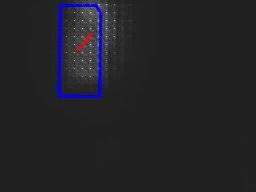}
	\caption{Left: An example input image. Right: The output of the layer detecting $40\times\SI{40}{\mm}$ aluminium profiles. The ratio of activations inside the area of the object, displayed in blue,  divided by the sum of all activations and the size of the blue area produces the IoU. The Euclidean distance error between weighted center of activations and ground truth annotation is displayed in red.}
	\label{fig:func_iou}
\end{figure}

\begin{figure}[b]
	\centering
	\includegraphics[width=1\linewidth]{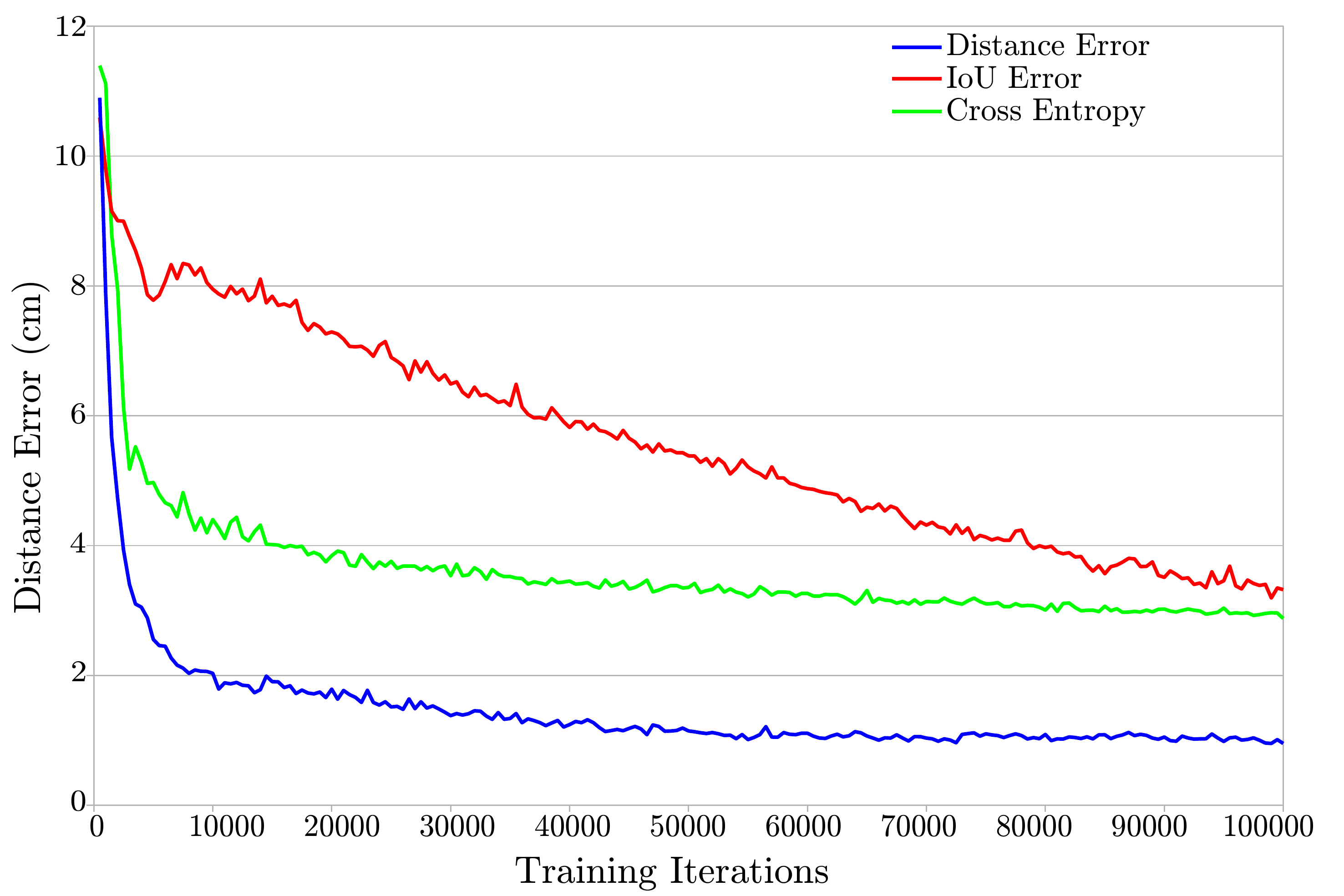}
	\caption{Training directly on the distance error metric (blue) leads to faster convergence and lower error than training on cross entropy (green) or IoU (red). Averaged from 20 runs with equal hyperparameters.}
	\label{fig:one_error_funcs}
\end{figure}

\begin{figure}[tb]
	\centering
	\includegraphics[width=0.32\linewidth]{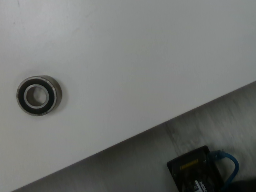} \includegraphics[width=0.32\linewidth]{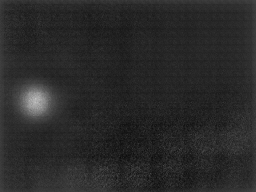} \includegraphics[width=0.32\linewidth]{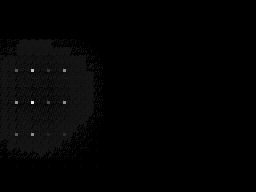} 
	\caption{Left: The input image leading to the activations shown right. Middle: Activations on the \enquote{Bearing} layer for a network trained on IoU. Right: Activations when trained on distance error alone. The object position is predicted accurately by constructing the weighted average of the activations, however, the majority of the filter weights are zero, and most of the activations are not close to or within the object. Note that the size of the activated pixels was increased for better visibility.}
	\label{fig:sparse_activation_distance}
\end{figure}

Commonly, object detection or semantic segmentation networks are trained on minimizing the cross entropy between the network output and the labeled ground truth. Training to maximize the IoU between the predicted and actual bounding boxes is not possible out of the box, as this function is not differentiable; however, one can use an approximate IoU measure instead, which~\cite{rahman2016optimizing} argue converges more quickly. Note that in this paper we are mainly minimizing the \emph{IoU Error}, defined as $1 - IoU$. Figure~\ref{fig:func_iou} visualizes how the IoU measure is calculated from a prediction and approximate segmentation data, obtained by a conventional object detection approach.
In case the prediction and ground truth do not overlap, the IoU does not produce a usable gradient and training can get stuck. The Euclidean distance between prediction and label, however, is always well defined and produces a slope a gradient descent algorithm can follow.
Because of this, and because the measure of quality for a picking operation will ultimately be a function of the distance between predicted and actual position of the object, one might train the network on this measure immediately. The object prediction is produced by obtaining the weighted average of activations on the corresponding output layer, as indicated in red in Figure~\ref{fig:func_iou}. Minimizing this \emph{distance error} lets the network converge faster and to a lower error, as Figure~\ref{fig:one_error_funcs} demonstrates. However, low-frequency filters appear to dominate the output, in turn leading to activations being spread over the greater part of the image, as opposed to tightly localized around the desired position. Figure~\ref{fig:sparse_activation_distance} shows this behavior. The distinction between multiple objects on the same layer using clustering is therefore impossible. As the IoU measure produces strongly localized activations, but the distance error converges much faster, we hypothesize that a combination of both can produce superior results.

An inherent weakness of the \emph{distance error} in pixels or centimeters is its lack of expressiveness. Although it does describe the visual offset between prediction and label, it fails to capture the quality of this prediction: While halving an error from e.g. \SI{0.8}{\cm} to \SI{0.4}{\cm} will not result in any increased chance of a successful pick-up operation, halving the error from \SI{2}{\cm} to \SI{1}{\cm} will greatly increase the probability. The relation between a successful pickup and the distance error is therefore not linear. During empirical testing of the picking operation, we observed that for small objects, a distance error of \SI{1}{\cm} results in a failed grasp $50\percent$ of the time; for a distance offset of \SI{3}{\cm}, the probability of failure rises to $90\percent$. We thus model the \emph{pickup error} by applying a negative exponent power function to the error in centimeters, coinciding in $E_{Distance}(0) = E_{Pickup}(0) = 0$, and $E_{Pickup}$ asymptotically approaching $1$ for $E_{Distance}\to\infty$, resulting in Equation~\ref{eq:pickup_error}.

\begin{equation}
\label{eq:pickup_error}
E_{Pickup} = \frac{-1}{(E_{Distance})^2+1} + 1
\end{equation}

\begin{figure}[b]
	\centering
	\includegraphics[width=\linewidth]{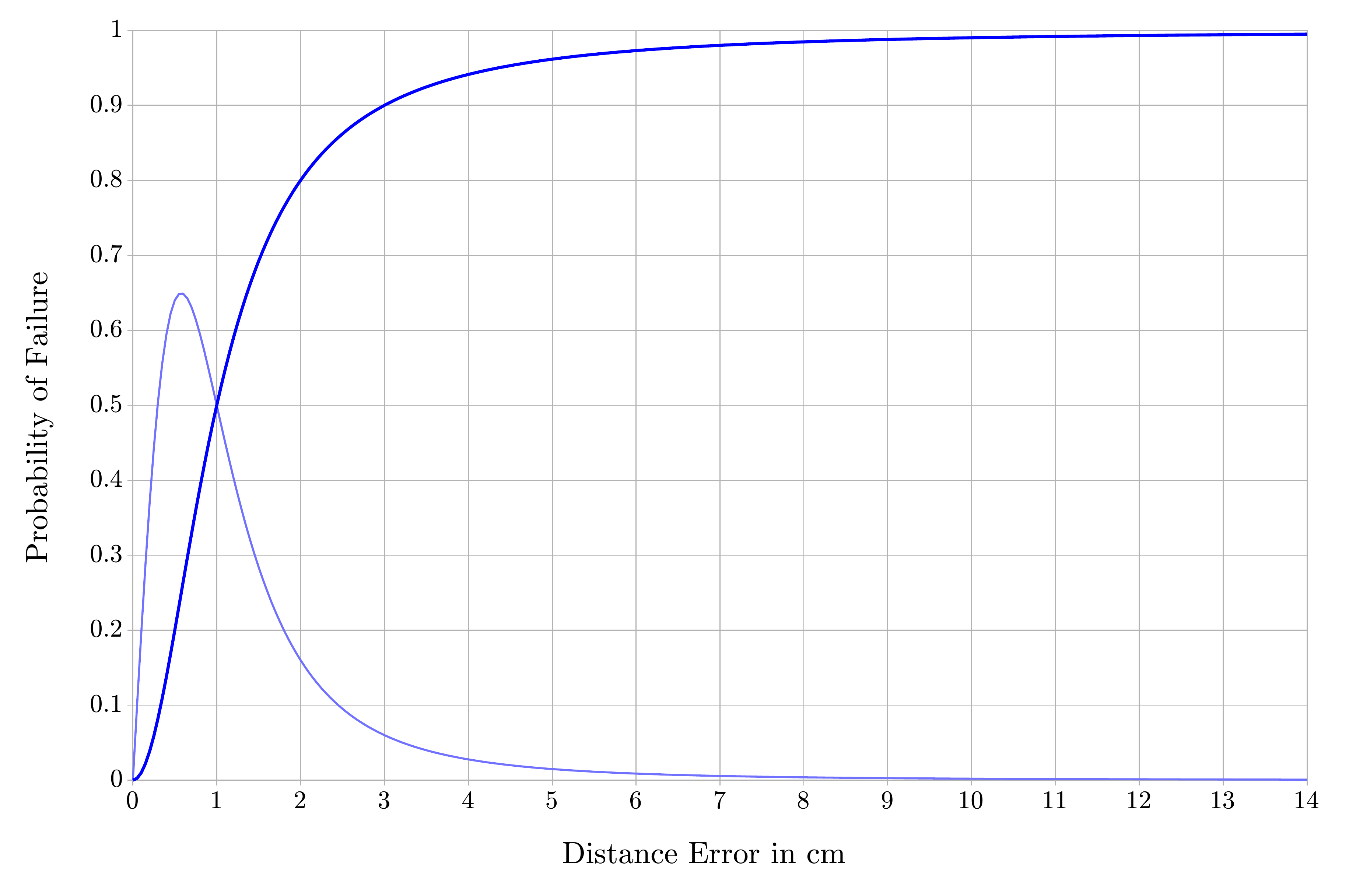}
	\caption{The probability of a failed pickup (blue) and its derivative (light blue) plotted in a range of 0 to \SI{14}{\cm} of distance error.}
	\label{fig:pickup_error_func}
\end{figure}

Figure~\ref{fig:pickup_error_func} plots the distance error metric versus the pickup error in the usually occurring range. Using the pickup error metric readily scales the distance error in the same range as the IoU, between $0$ and $1$.
Early results showed that training on pickup error directly was unsuccessful, due to the gradient of the function being too low for higher error values.
However, we will use the pickup error and the \emph{pickup rate}, defined as $1 - E_{Pickup}$, as an accurate scaling method when comparing or adding pickup and IoU errors.

\section{Experiments}
\label{sec:exp}
To measure the convergence speed and resulting quality of employing different error functions and combinations thereof, each setup was trained for 100000 steps ($\approx 67$ epochs) with a batch size of 8. An Adam optimizer~\cite{kingma2014adam} was used to minimize the loss term, with a learning rate of \num{1e-5}, $\beta_1=0.9$, $\beta_2=0.999$, and $\epsilon=\num{1e-08}$.
Initial tests showed that all methods tested could train and converge with these values, so the hyperparameters for the neural networks to be trained were frozen at these values.
In order to achieve statistical significance, 20 networks were trained for every method. A full training run with 100000 iterations takes about 10 hours to complete on an NVIDIA Titan X graphics card, regardless which of the methods we present is used.
Every 500 iterations during training, the networks were tested on the entire validation set to get an accurate estimate of their performance on unknown data. This data forms the basis for all results to be presented.

\subsection{Dataset}
\begin{figure}[tb]
	\centering
	\includegraphics[width=0.32\linewidth]{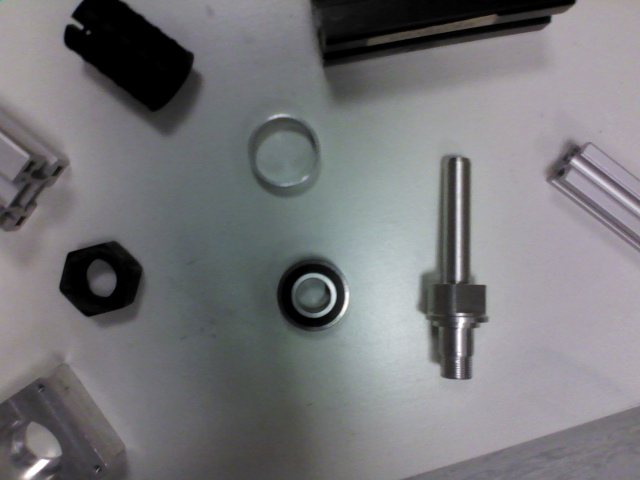} \includegraphics[width=0.32\linewidth]{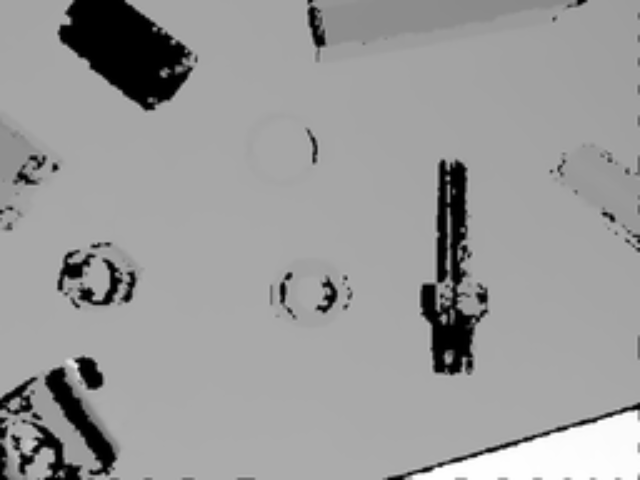} \includegraphics[width=0.32\linewidth]{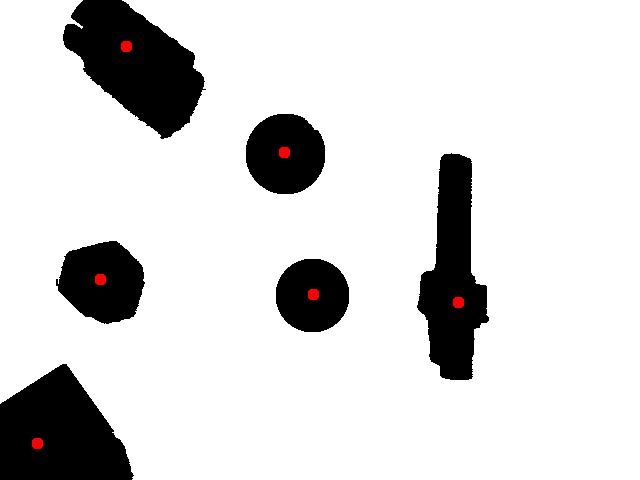}
	\caption{Left: An example image from the dataset, displaying multiple {\rc} objects. Center: The corresponding depth image. Right: Imperfect segmentation mask obtained by the current {\rc} software. Manually labeled pickup positions marked in red.}
	\label{fig:dataset_example}
\end{figure}

Employing the {\rc} challenge as a Factory of the Future simulation, the up-to-date version of team \emph{smARTLab@Work}'s previously world-cup winning~\cite{broecker2014winning} hard- and software was used to produce a dataset. More than $35000$ RGBD 3D images were taken with an Intel SR300 3D camera, and over $14000$ images were manually labeled for object detection.
Alternative RGBD benchmark datasets containing {\rc} objects are scarse. 
Distinct training and evaluation sets were recorded, mimicking the Industry~4.0 requirements such that both are comparatively small with about 12000 images for training, and 2000 images for validation. Each of the 13 {\rc} object types is recorded from multiple angles, placed on platforms as defined in the official rule book\footnote{Available on \tt{http://www.robocupatwork.org/}}. Figure~\ref{fig:dataset_example} shows an example.
The training and validation sets were manually labeled with the center of gravity, or \enquote{pickup point}, of each object. The dataset can be retrieved from {\footnotesize \tt{https://airesearch.de/ObjectDetection@Work/}.}

\section{Results}
\label{sec:res}
The results of the experiments conducted are split into analyzing how and which weights are derived by the available methods to weight error functions, and their respective performance regarding convergence speed and error reduction.

\subsection{Optimal Weight Learning} \label{sec:res:owl}
In order to compare \emph{KGC-weighting} to the weighting learned by the auxiliary task network \auxnet, we are testing both methods to empirically evaluate their respectively derived combinations of the distance error and IoU error measures.
To produce a valid \emph{KGC-weighting}, scaling the error functions between 0 and 1 is required. The IoU error is produced in this range by design, but the pixel distance error may range from 0 to 320 pixels, i.e. the diagonal of the image. In practice however, distances over 70 pixels do not occur. As a compromise, we divide the pixel distance by 100 in order to scale it in the required range. Tests using the pickup error as scaled distance error term during training were unsuccessful, evidently due to the lack of slope of the error function throughout most of its range.

Training the network on the \emph{KGC-weighted} downscaled pixel distance and IoU posed another challenge, as frequently, the network would irrecoverably die out, most certainly due to the ReLU activation functions getting stuck with negative weights, a well known problem with ReLU activation function and steep gradients~\cite{maas2013rectifier}. In order to get 20 successful training runs, 30 networks had to be trained, as one third of them did not converge. 
A contributing factor to the dying ReLUs appears to be the extreme weights derived from the task variance. Dividing by the very low values of initial IoU variance leads to extreme weights, which in turn can lead to a gradient explosion.
Adding an epsilon of \num{1e-3} to both variances mitigated the neuron decay, however, with a detrimental effect on the prediction performance. We call this method \emph{$KGC_{+\epsilon}$-weighting}.

\begin{figure}[tb]
	\centering
		\begin{minipage}{0.49\linewidth}\centering $KGC$ \end{minipage}\begin{minipage}{0.49\linewidth}\centering $KGC_{+\epsilon}$ \end{minipage}\\
		\includegraphics[width=0.49\linewidth]{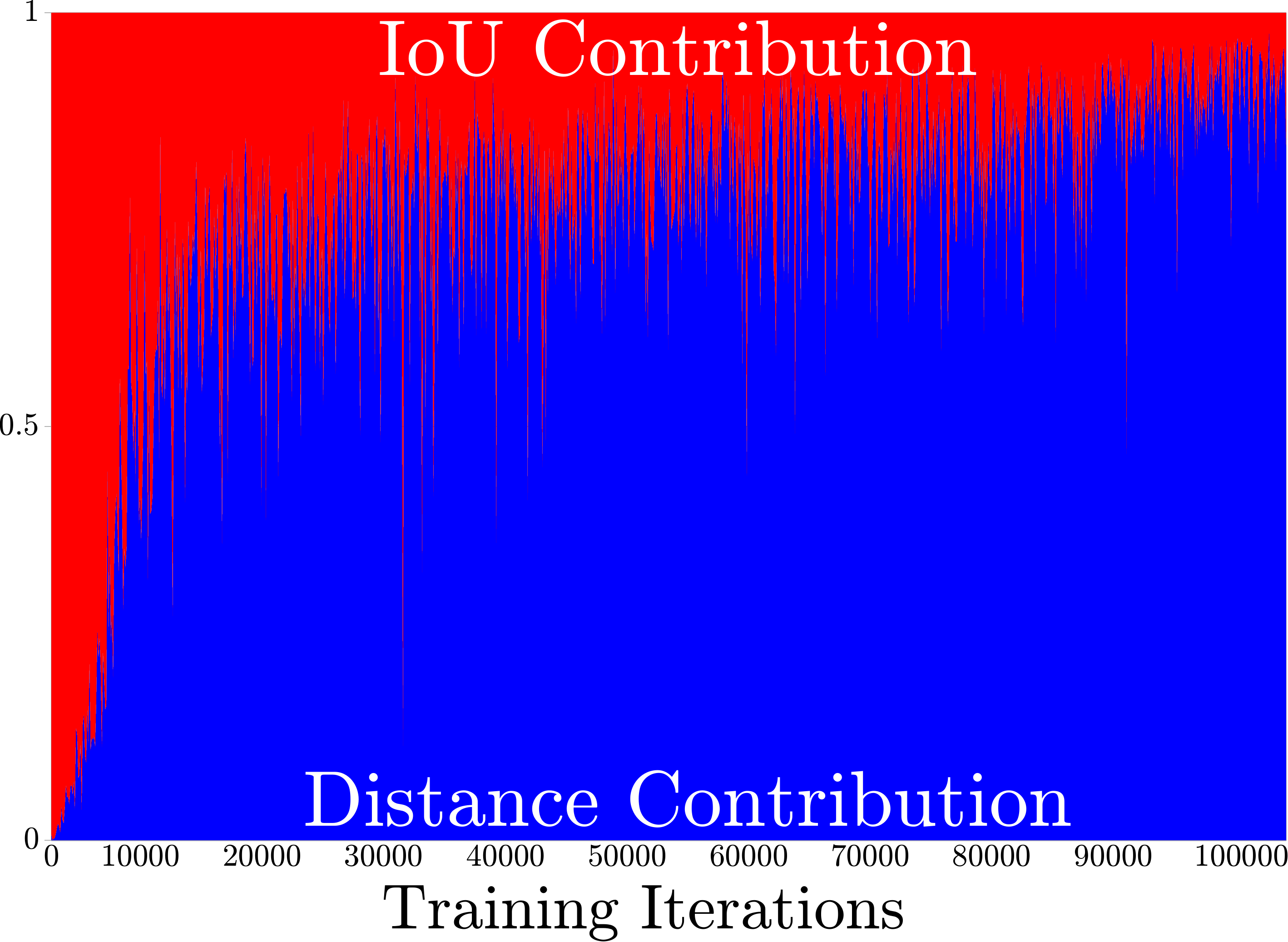}
		\includegraphics[width=0.49\linewidth]{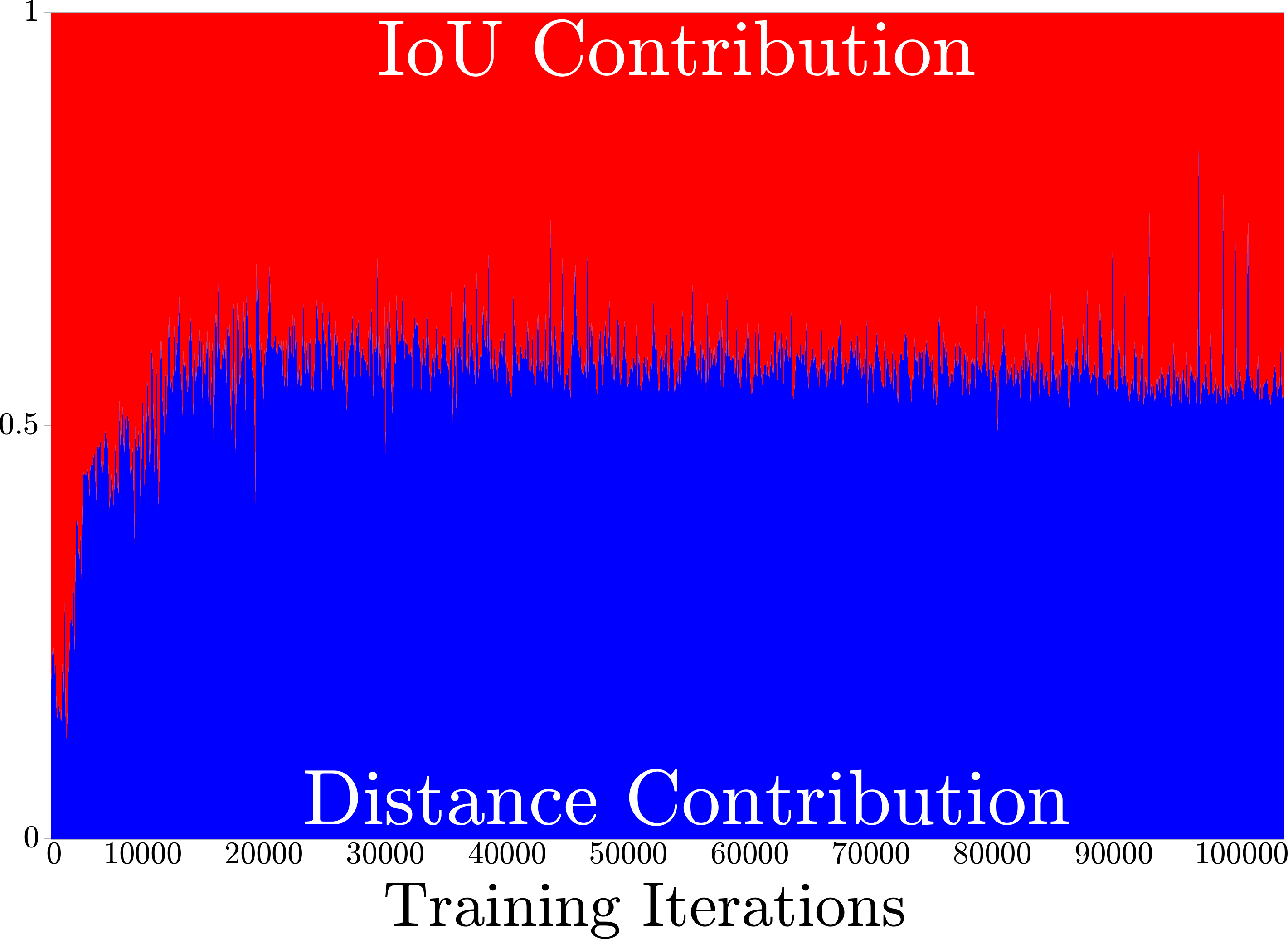}
		\begin{minipage}{1\linewidth}\vspace{0.2cm}\end{minipage}
		\begin{minipage}{0.49\linewidth}\centering $KGC_{\nicefrac{}{Mean}}$ \end{minipage}\begin{minipage}{0.49\linewidth}\centering {\auxnet} \end{minipage}\\
		\includegraphics[width=0.49\linewidth]{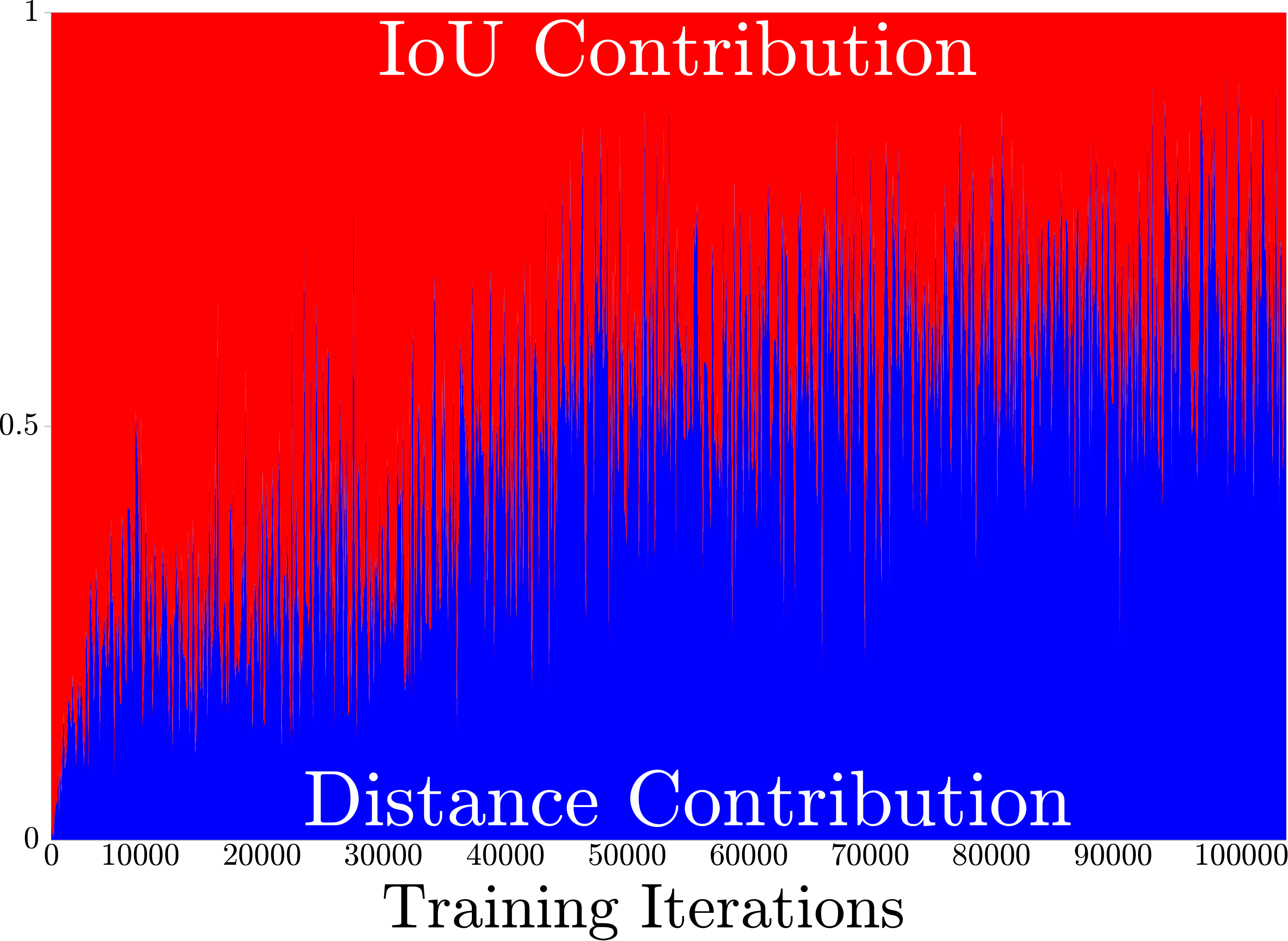}
		\includegraphics[width=0.49\linewidth]{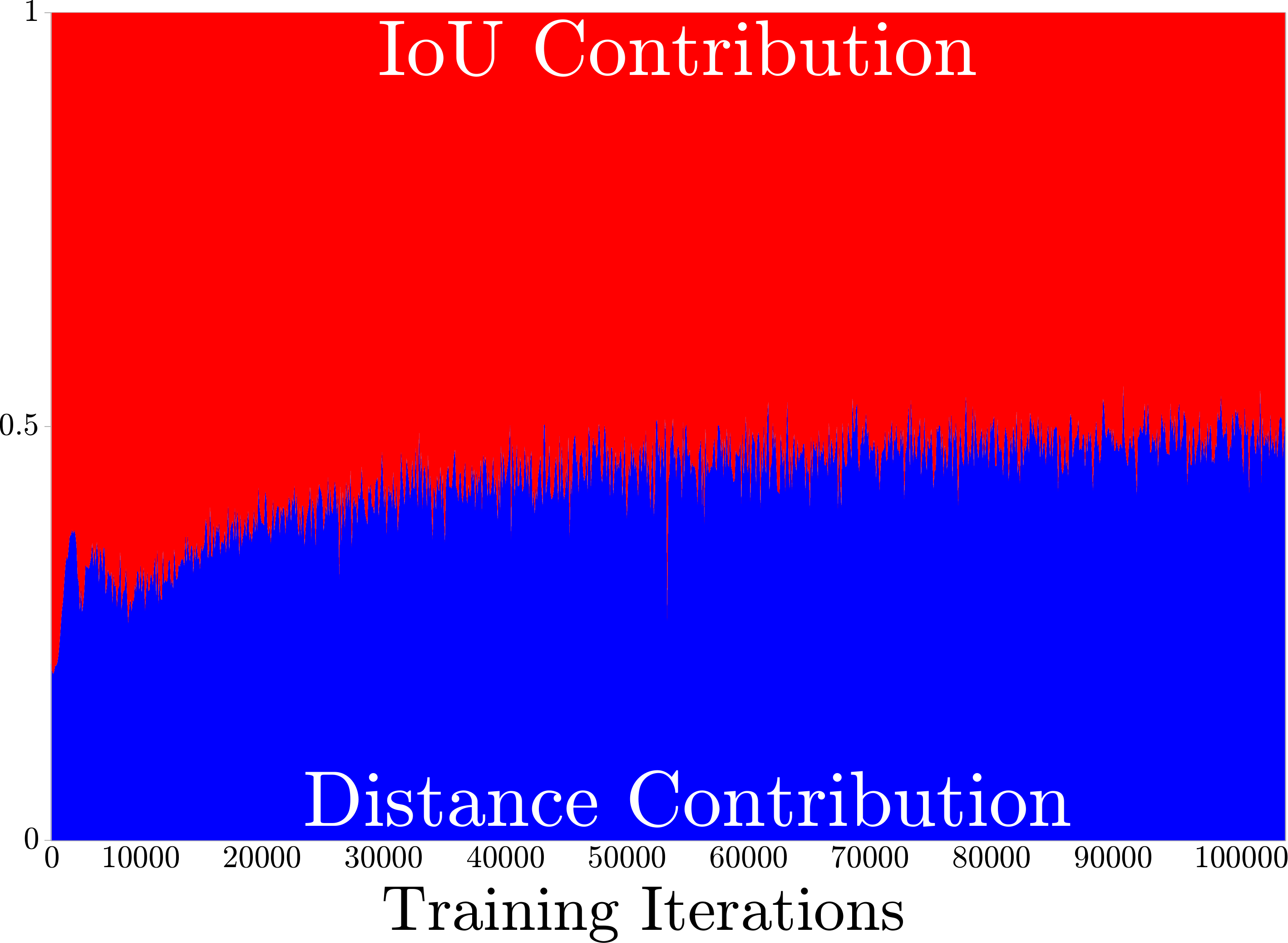}

	\caption{Normalized and inverted weights derived from the different methods. Positive contribution to distance error is shown in blue, contribution to the IoU error is shown in red. All $KGC$-measures start off with training almost exclusively on IoU. Adding an epsilon balances the weights more. {\auxnet} features more balanced weights from the start, explaining the fast early convergence. The weights appear to be more stable than $KGC$'s.}
	\label{fig:kgc_weights_chart}
\end{figure}

\begin{figure}[tb]
	\centering
	\includegraphics[width=1\linewidth]{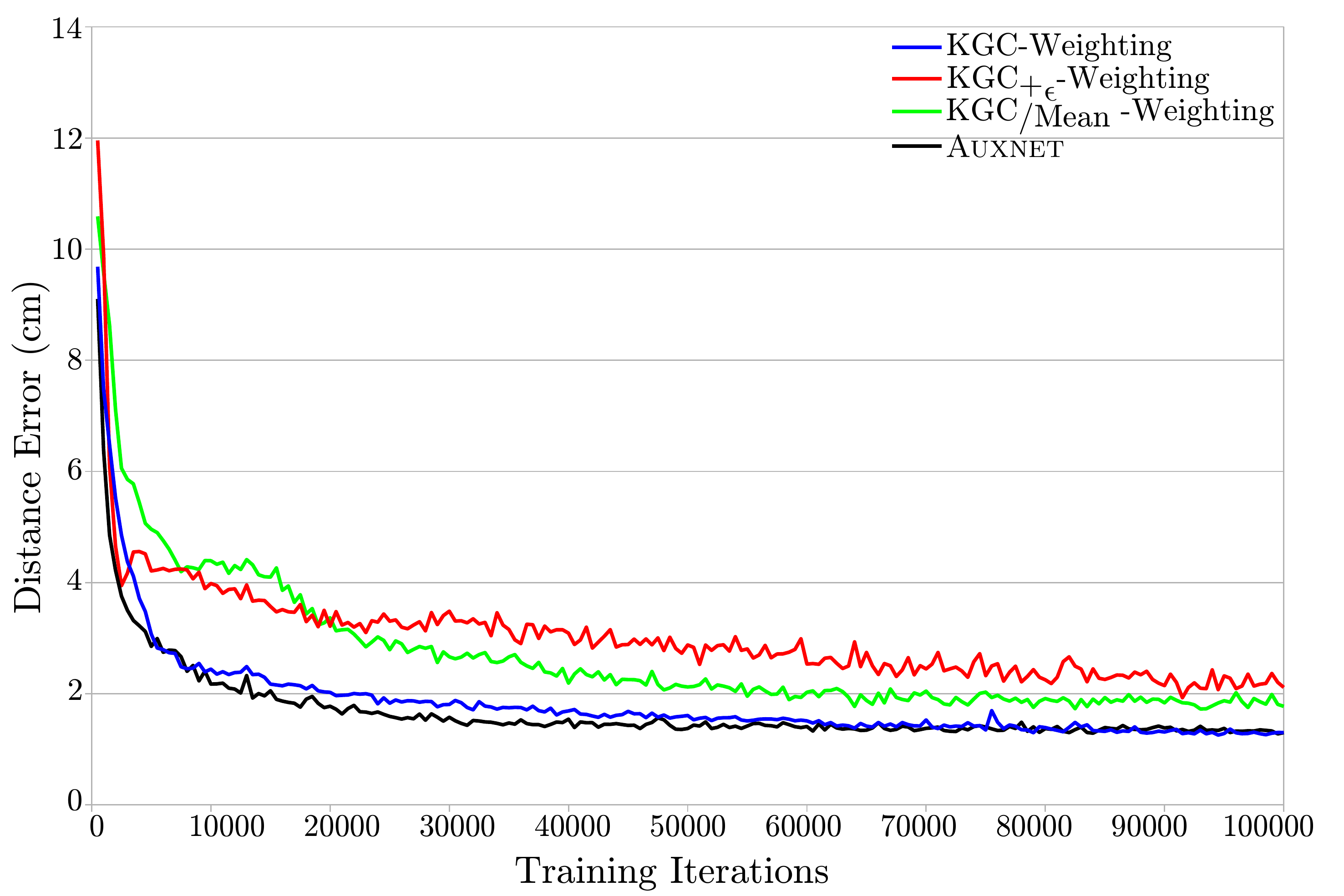} \\
	\includegraphics[width=1\linewidth]{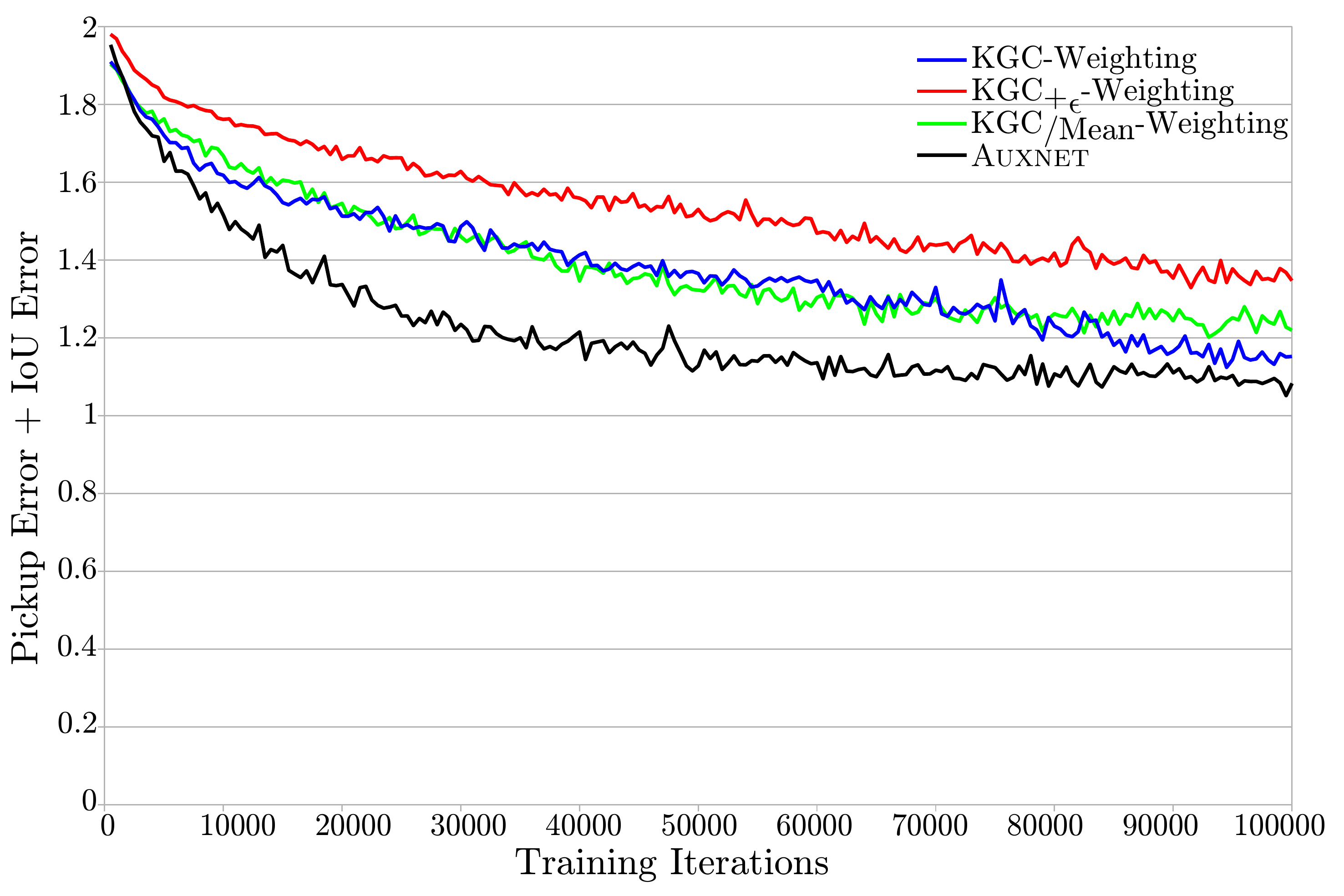} \\
	\caption{Charts comparing the reduction of different error measures for (blue) \emph{KGC-weighting}, (red) \emph{$KGC_{+\epsilon}$-weighting}, (green) \emph{$KGC_{\nicefrac{}{Mean}}$-weighting} and (black) weighting by {\auxnet}. The first chart plots the distance error in cm, the second shows the reduction of an added pickup- and IoU error measure. Charts are averaged from all 20 runs and obtained on the entire evaluation set during training.}
	\label{fig:kgc_charts}
\end{figure}

Another way to combat exploding gradients is to divide the error variance by the error mean, assuming that an error function with a higher error in general will also feature a higher absolute variance. This \emph{$KGC_{\nicefrac{}{Mean}}$-weighting} performs comparable to regular \emph{KGC-weighting}, and does not require explicit scaling of the error functions to a 0 to 1 range. It also appears to generate more sensible weights, as the network converged in all 20 out of 20 runs. Table~\ref{tab:weights_Learned} shows the extreme values of the learned weights. Figure~\ref{fig:kgc_weights_chart} presents the weight history over a full training episode. For these charts, the weights were normalized and inverted, so a larger area means a higher contribution of the weight to the loss term. As $KGC_{\nicefrac{}{Mean}}$ and {\auxnet} are trained on the raw pixel distance rather than a scaled distance between 0 and 1, their weights are scaled accordingly by 100.
Figure~\ref{fig:kgc_charts} shows the effect of the different weight learning methods on error reduction during training.

\begin{table}[tb]
\centering
\setlength\tabcolsep{4.5pt} 
\begin{tabular}{|l|c|c|}
\hline
Learning method & $W_{Distance}$ & $W_{IoU}$ \\
\hline
\emph{KGC-weighting} & &  \\
\hspace{2em} min & \num{6.14e-6} & \num{2.26E-07} \\
\hspace{2em} max & \num{8.57e-2} & \num{2.18E-01} \\
\hline
\emph{$KGC_{+\epsilon}$-weighting} & &\\
\hspace{2em} min & \num{2.25E-01} & \num{1.00E-03} \\
\hspace{2em} max & \num{8.29E+02} & \num{2.19E-01} \\
\hline
\emph{$KGC_{\nicefrac{}{Mean}}$-weighting} & & \\
\hspace{2em} min & \num{7.98E-01} & \num{3.98E-05} \\
\hspace{2em} max & \num{6.44E+01} & \num{2.92E-02} \\
\hline
{\auxnet} & & \\
\hspace{2em} min & \num{2.71E-04} & \num{4.95E-03} \\
\hspace{2em} max & \num{7.76E-03} & \num{5.17E-02} \\
\hline
\end{tabular}

\caption{Extreme values for the error function weights learned by different methods. \emph{KGC-weighting} can produce the most extreme gradients, dividing the losses by the smallest weights. {\auxnet} appears to provide the most stable weights.}
\label{tab:weights_Learned}
\end{table}

\subsection{Convergence Speed and Accuracy}
\begin{figure}[tb]
	\centering
		\includegraphics[width=0.19\linewidth]{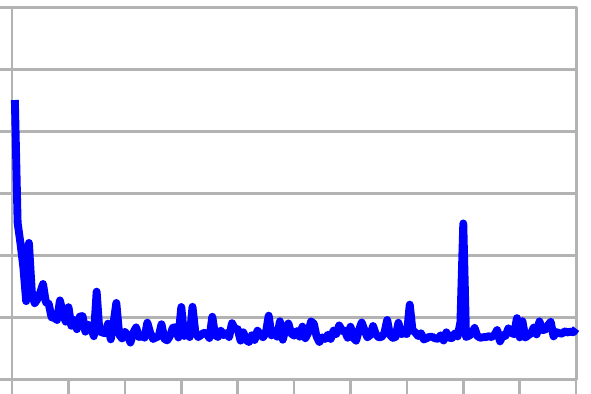}\includegraphics[width=0.19\linewidth]{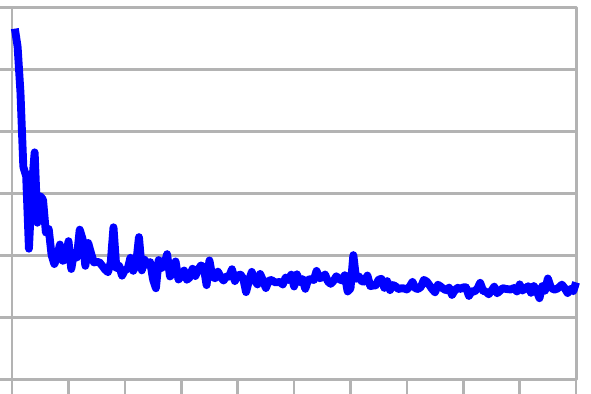}\includegraphics[width=0.19\linewidth]{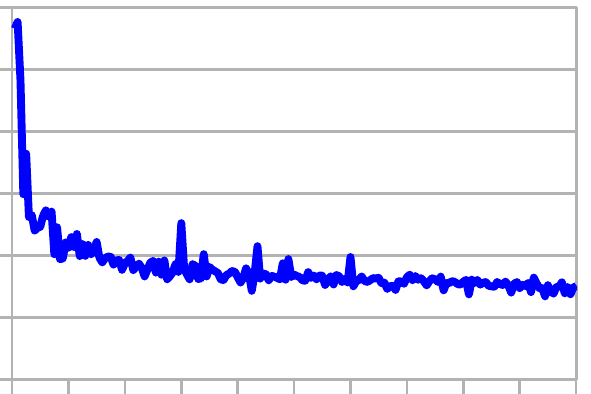}\includegraphics[width=0.19\linewidth]{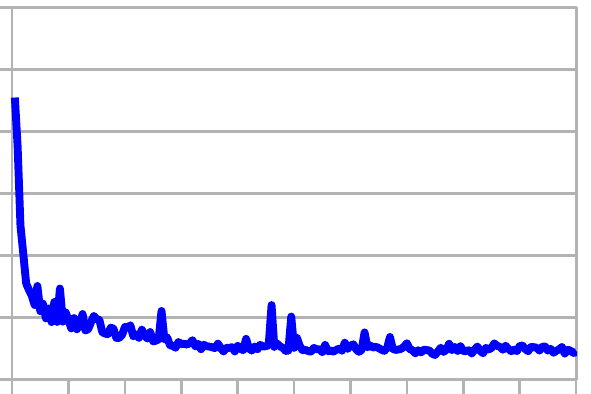}\includegraphics[width=0.19\linewidth]{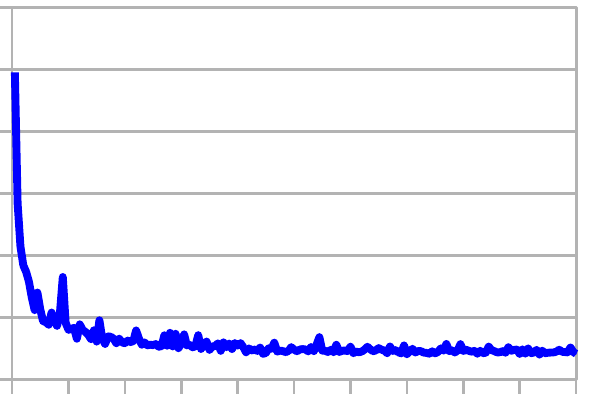}
		\includegraphics[width=0.19\linewidth]{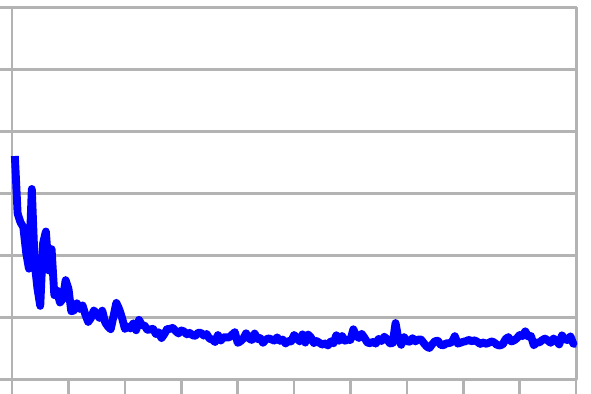}\includegraphics[width=0.19\linewidth]{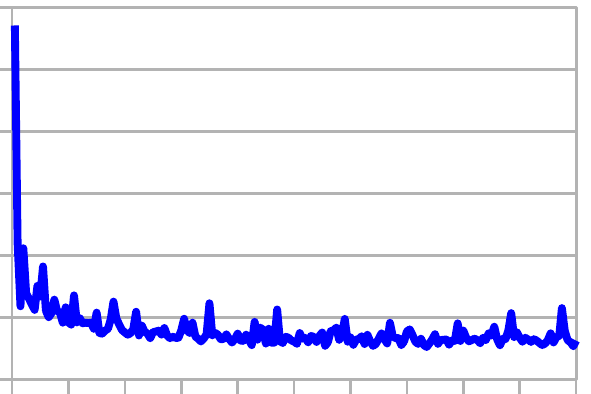}\includegraphics[width=0.19\linewidth]{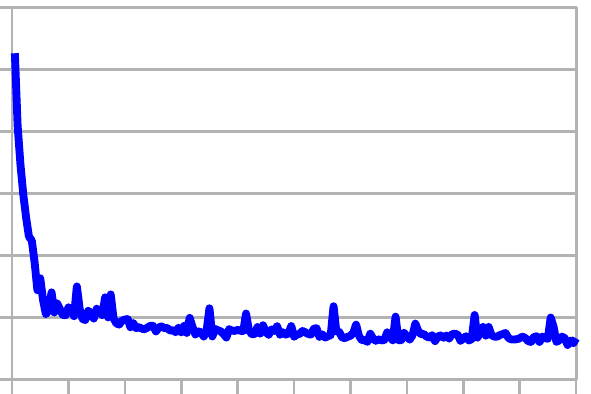}\includegraphics[width=0.19\linewidth]{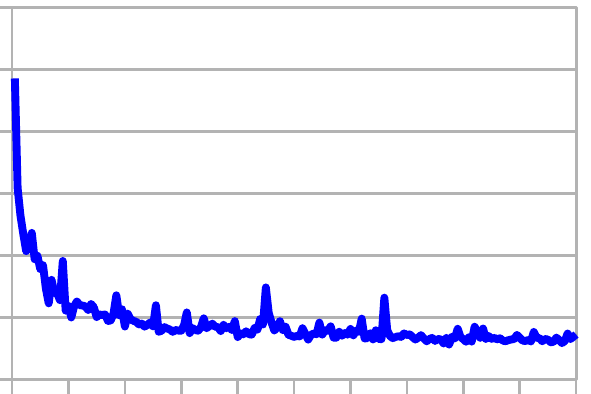}\includegraphics[width=0.19\linewidth]{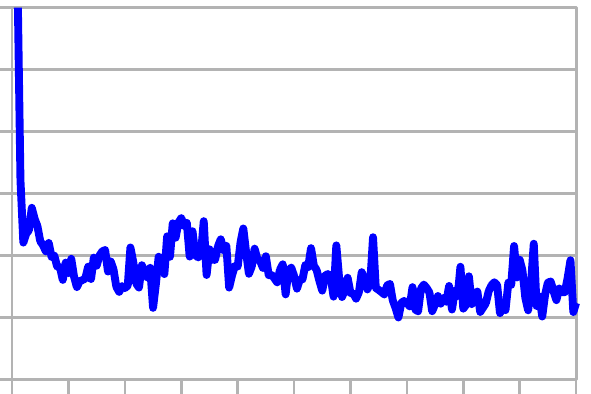}
		\includegraphics[width=0.19\linewidth]{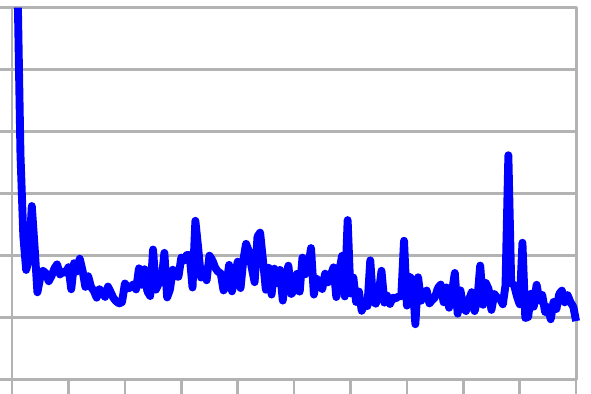}\includegraphics[width=0.19\linewidth]{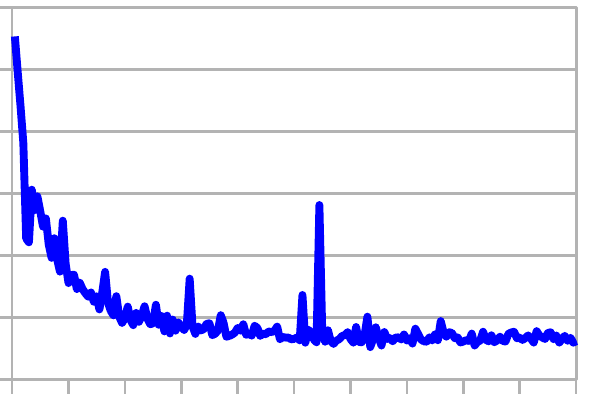}\includegraphics[width=0.19\linewidth]{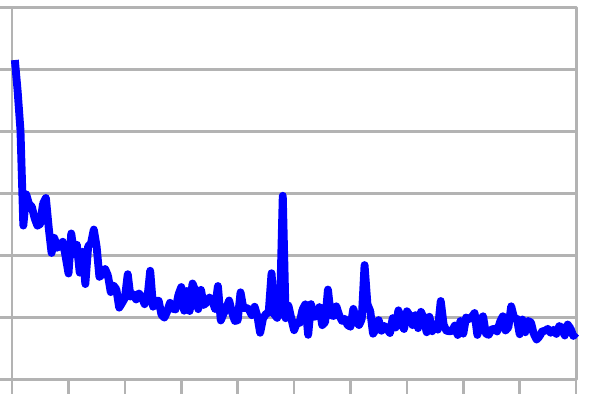}\includegraphics[width=0.19\linewidth]{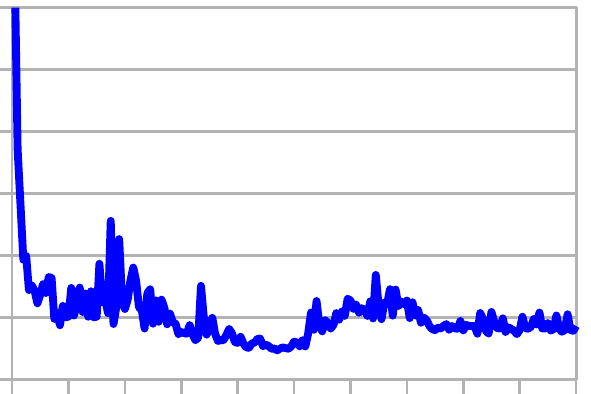}\includegraphics[width=0.19\linewidth]{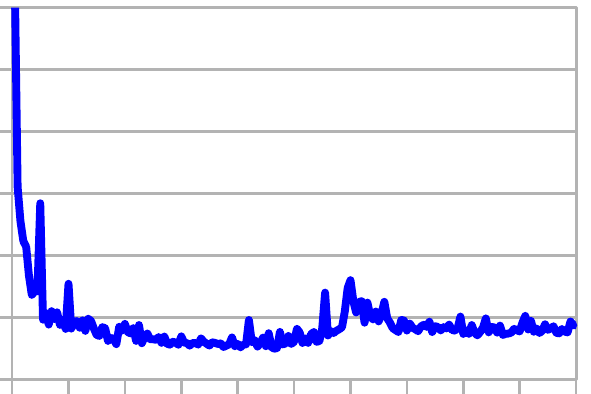}
	\caption{All test runs were examined for proper convergence. $15$ randomly selected, successfully converging charts are shown here for convenience.}
	\label{fig:many_charts}
\end{figure}

\begin{figure*}[tb]
	\centering
	\includegraphics[width=0.32\linewidth]{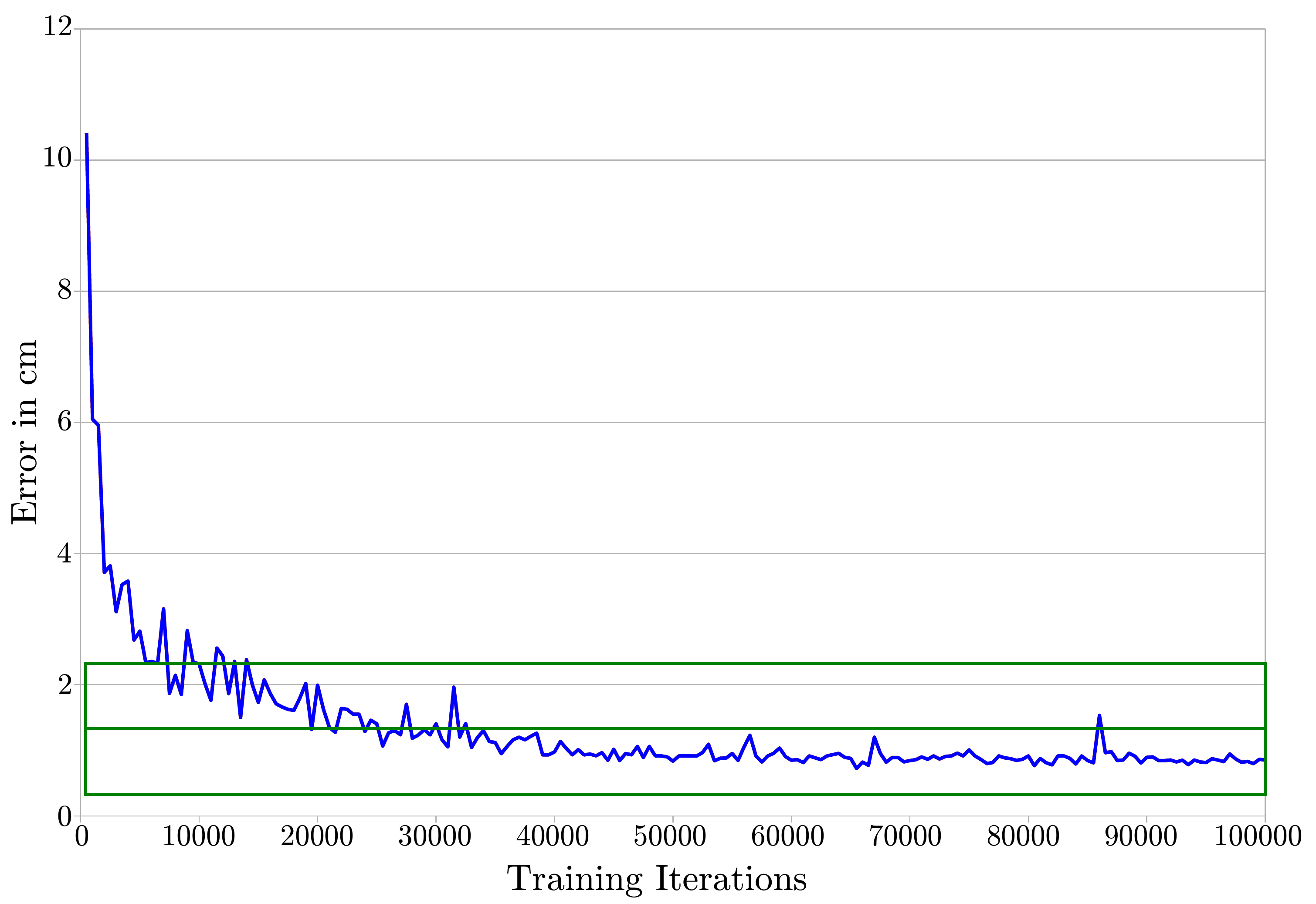}\includegraphics[width=0.32\linewidth]{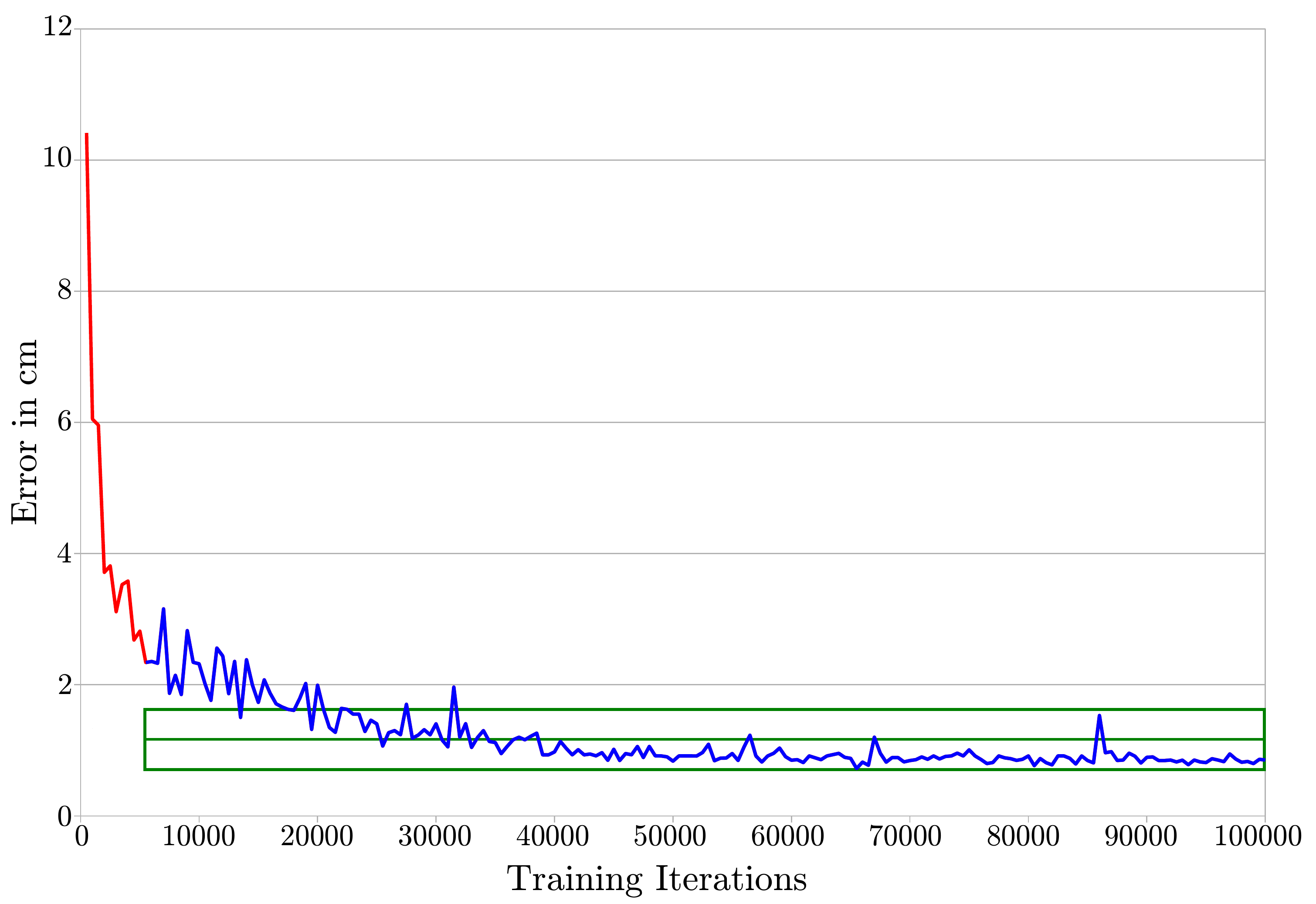}\includegraphics[width=0.32\linewidth]{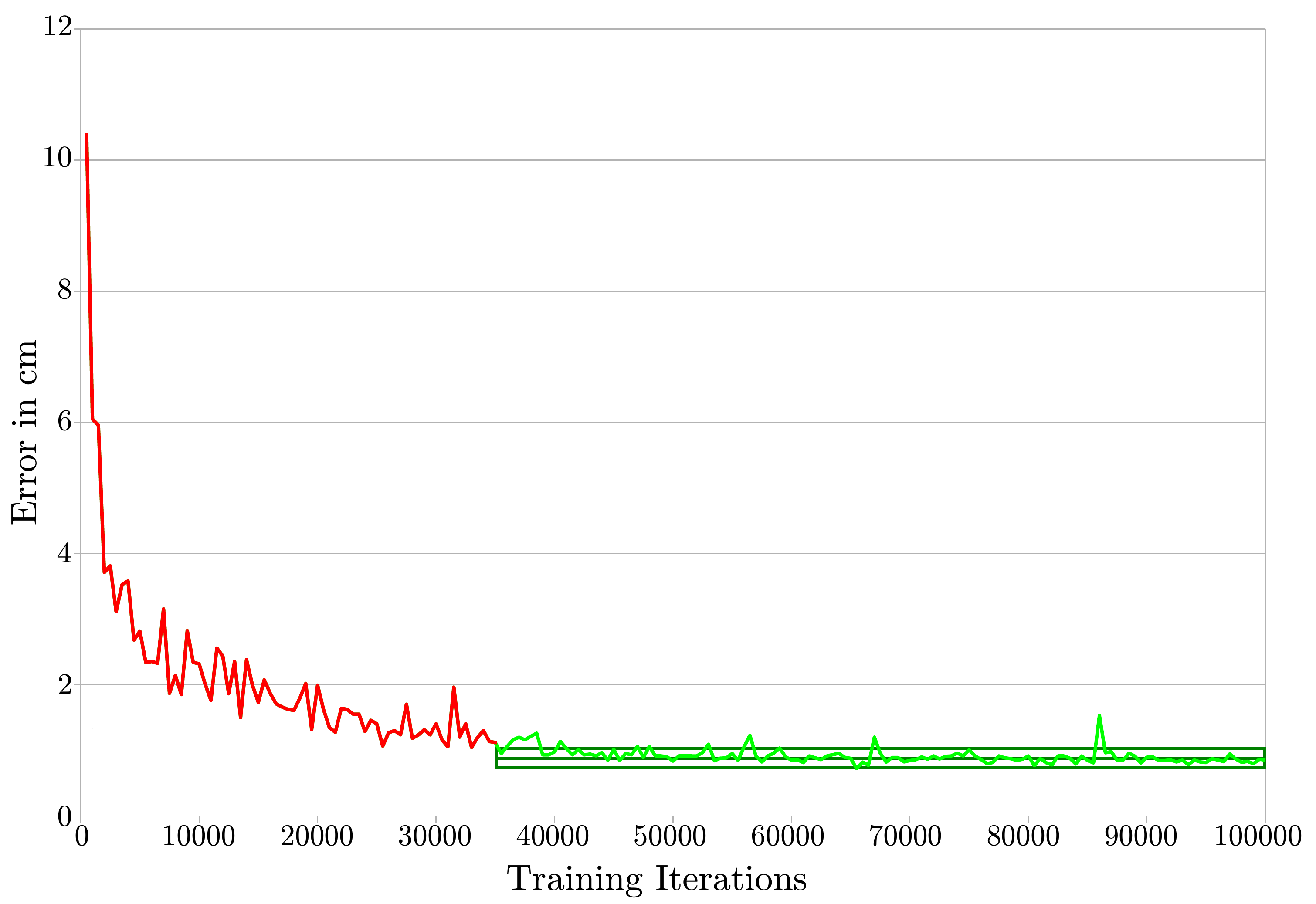}
	\caption{The convergence point determining process in detail. Left: An acceptance window (green) is created from the average error and the standard deviation. Middle: While rejecting points (red), the acceptance window shrinks. Right: The leftmost point within the acceptance window becomes the convergence point. For the amount of noise present in our data, a window of $\pm\nicefrac{\sigma}{2}$ produced best results.}
	\label{fig:howto_converge}
\end{figure*}

To compare the trained classifiers, convergence speed and error after convergence are measured separately, as in different scenarios, users may value time and quality constraints differently.
Using dropout regularization effectively prevented our networks from overtraining; however, it also introduced fluctuations of the error value after convergence.
These fluctuations prohibit the use of an epsilon threshold on the discrete derivative of the error as convergence measure, as the relation between selected epsilon and the resulting convergence point is highly nonlinear, and varies for networks trained on different error functions.
For our convergence point determination method, we instead assume a normal distribution of errors around the mean error after convergence. Further, we assume that the longer we train a network, the more it converges, i.e., that it does not diverge and also that it always reaches convergence within 100000 iterations. Visual analysis of the error charts supports these assumptions, as hinted by $15$ randomly selected charts all converging, shown in Figure~\ref{fig:many_charts}.
We then iteratively remove the earliest data points that do not fit a normal distribution around the mean error with regards to the standard deviation, and shrink the acceptance window to the newly obtained mean and standard deviation. Figure~\ref{fig:howto_converge} illustrates this process. The first entry of the remaining data is considered to be the convergence point, and the average of the remaining data can be considered the average error after convergence.

\begin{figure}[tb]
	\centering
	\includegraphics[width=1\linewidth]{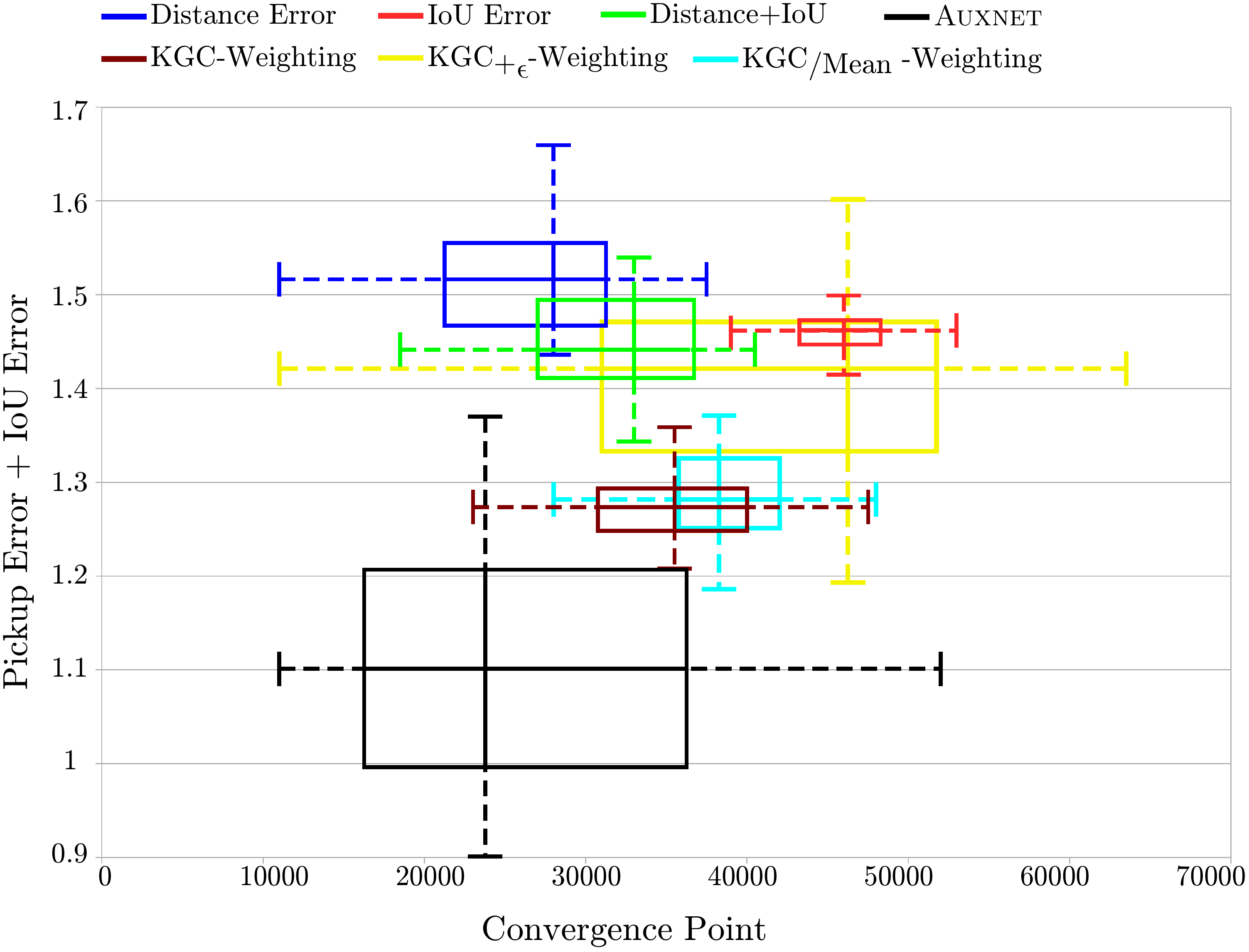}
	\caption{A double box plot showing the median and quartiles from twenty networks trained on each error function or combination thereof. Extents along the X-Axis describe variation in convergence speed, while the Y-Axis shows the summed \emph{pickup}- and IoU error. The median convergence and error measures for {\auxnet} are significantly better than for any compared method.}
	\label{fig:double_box_plot}
\end{figure}

Figure~\ref{fig:double_box_plot} displays the convergence speed and resulting error after convergence in a double box plot. This plot shows that training on the distance error alone yields the highest error, although it trains comparatively fast. Training on IoU error alone converges very slowly, but to a lower error compared to the distance error metric. This is due to the fact that training on IoU error reduces both the IoU and distance error, but training on the distance error does not lower the IoU error. Training on a simple addition of distance error and IoU results in reduced error compared to either. Both \emph{KGC-weighting} and \emph{$KGC_{\nicefrac{}{Mean}}$-weighting} significantly reduce the combined error, but do not improve the convergence time.
While {\auxnet} features a higher deviation in both convergence speed and quality, the majority of the runs perform significantly better than any other combination method, reducing the average summed error by $15.85\percent$ compared to \emph{KGC-weighting}; it also proved to be more stable, as all 20 test runs converged. {\auxnet} manages to weight the error terms such that the network can converge on average in $26800$ iterations, much faster than \emph{KGC-weighting} with an average of $34850$ iterations -- reducing it by nearly $25\percent$, or saving about one hour of training on an Nvidia Titan X. Table~\ref{tab:improvements} shows the relative improvement of {\auxnet} compared to just training on the distance error, IoU error, on a simple addition of IoU and Distance Error, and \emph{KGC-weighting}.

\begin{table}[tb]
\centering
\setlength\tabcolsep{4.5pt} 
\begin{tabular}{|l|c|c|c|c|}
\hline
{\auxnet} vs. & Convergence Time & Pickup Rate & IoU \\
\hline
Distance Error       & ~$+0.48\percent$ & $+11.83\percent$ & $+35.31\percent$\\
IoU Error            & $-42.48\percent$ & $+39.90\percent$ & ~$+3.31\percent$\\
Distance+IoU         & $-16.90\percent$ & $+21.53\percent$ & $+24.41\percent$\\
\emph{KGC-weighting} & $-24.53\percent$ & $+15.85\percent$ & $+10.48\percent$\\
\hline
\end{tabular}
\caption{Relative improvements of {\auxnet} over selected other methods. Data is averaged from 20 runs. {\auxnet} converges faster and produces a higher pickup rate and IoU than the other tested methods, an exception being the distance error converging marginally faster.}
\label{tab:improvements}
\end{table}

\begin{figure*}[tb]
	\centering
	\includegraphics[width=0.49\linewidth]{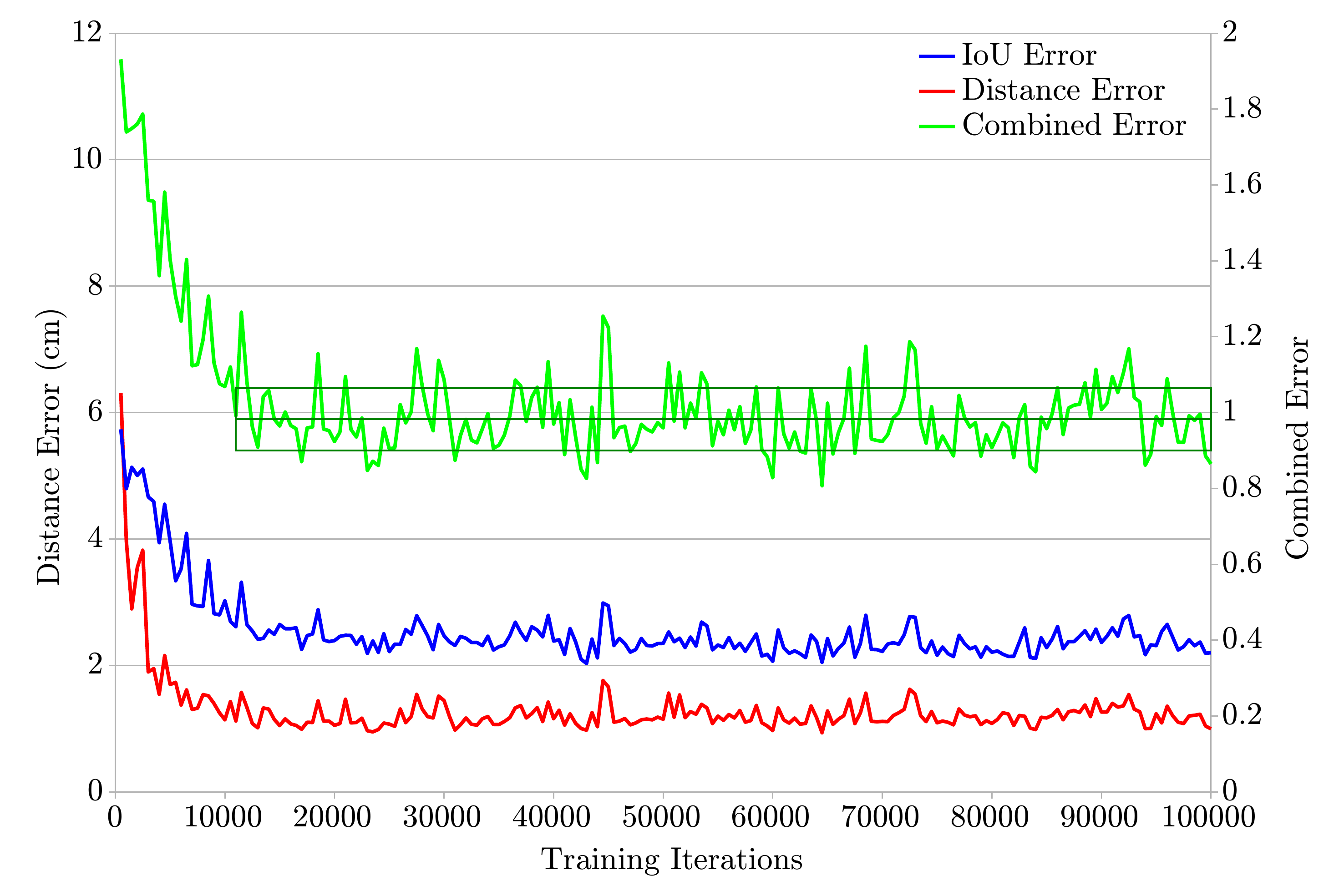}\includegraphics[width=0.49\linewidth]{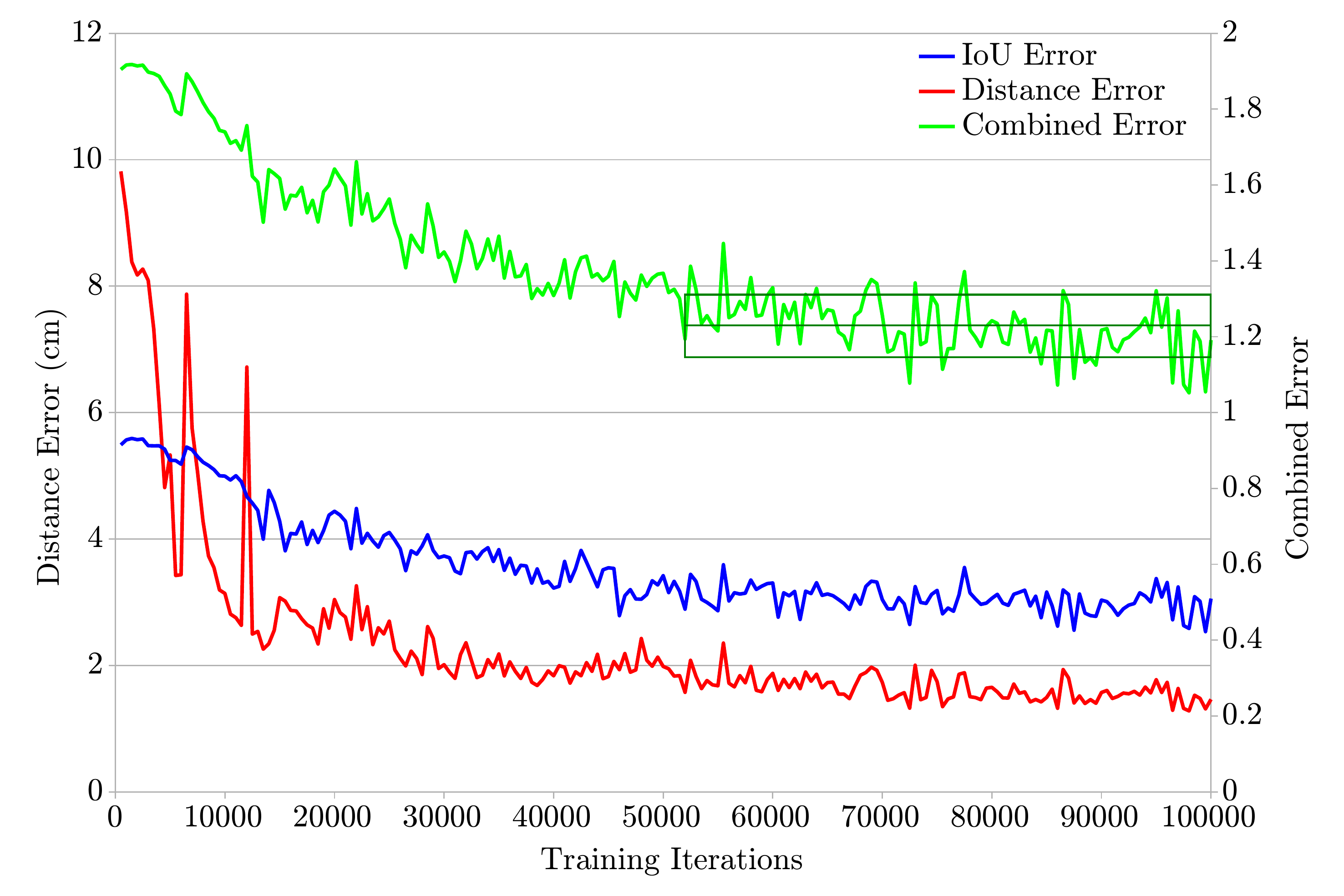}
	\caption{Convergence plots of the fastest (left) and slowest (right) converging networks trained using {\auxnet}. Distance error (red) is converted to a pickup error and added to the IoU error (blue) to form the combined error (green). The dark green box indicates the converged region, its average error, and standard deviation. The distance error plot on the right side shows that the network suffered traumatic experiences during early training, from which it could never fully recover, resulting in overall higher error and slower convergence.}
	\label{fig:comp_fast_slow}
\end{figure*}

The high variation of quality and convergence can be explained by analyzing the charts of the more slowly converging outliers. Figure~\ref{fig:comp_fast_slow} shows that these network experienced near-death scenarios, in which part of the network died out, from which they never fully recovered. It shall be noted that optimal weights for error functions will automatically also pose a higher threat of neuron death for the network, as optimal weights will produce steeper gradients, which in combination with Adam's learned momenta may lead to ReLU neurons getting stuck with negative activations and dying. An obvious solution may be converting the DCNN to leaky ReLUs. 

\subsection{Significance}
The distributions of error values and convergence points were tested against each other with a two-sample Kolmogorov-Smirnov (KS) test, using $\alpha = 0.05$, the null hypothesis being that the different sources can not be distinguished. All error distributions were statistically significantly different. Table~\ref{tab:significance} shows the results of the convergence tests. With distributions too similar or variations too high in either sample, some of the null hypotheses could not be rejected for the number of test runs.

\begin{table*}[tb]
\centering
\begin{tabular}{|l|c|c|c|c|c|c|c|}
\hline
Learning method & Distance Error & IoU Error & Distance+IoU & $KGC$ & $KGC_{+\epsilon}$ & $KGC_{\nicefrac{}{Mean}}$ & {\auxnet}\\
\hline
Distance Error            & \graycell & $\checkmark$ & $\checkmark$ & --- & $\checkmark$ & --- & --- \\
IoU Error                 & $\checkmark$ & \graycell & $\checkmark$ & $\checkmark$ & $\checkmark$ & $\checkmark$ & $\checkmark$\\
Distance+IoU              & $\checkmark$ & $\checkmark$ & \graycell & $\checkmark$ & $\checkmark$ & $\checkmark$ & $\checkmark$\\
$KGC$                     & --- & $\checkmark$ & $\checkmark$ & \graycell & $\checkmark$ & --- & $\checkmark$ \\
$KGC_{+\epsilon}$         & $\checkmark$ & $\checkmark$ & $\checkmark$ & $\checkmark$ & \graycell & $\checkmark$ & --- \\
$KGC_{\nicefrac{}{Mean}}$ & --- & $\checkmark$ & $\checkmark$ & --- & $\checkmark$ & \graycell & $\checkmark$\\
{\auxnet}                 & --- & $\checkmark$ & $\checkmark$ & $\checkmark$ & --- & $\checkmark$ & \graycell \\
\hline
\end{tabular}
\caption{Checkmarks indicating whether $H_0$ could be rejected by a two-sample KS-Test, $\alpha = 0.05$. All networks were compared with respect to their convergence. With few exceptions, most differences are significant, most importantly, between $KGC$ and {\auxnet}.}
\label{tab:significance}
\end{table*}

\section{Conclusion}
\label{sec:conc}
We have shown that error functions do not have to be independent in order to speed up convergence and produce lower error.
We have introduced the {\approach} approach, a method to learn a weighting between these error functions in a manner that improves convergence and the reduction of error even more. Repeated test runs have shown that improvements on state-of-the-art methods to derive optimal weightings are significant. In conclusion, object detection tasks can be improved in accuracy and training times by further constraining the network using a combination of available error metrics, and using approaches such as {\auxnet} to learn optimal weights between them.
The \emph{Factory of the Future} benefits especially from the faster convergence, enabling it to train object detectors for each new task in near real time.

Future work includes extending {\approach} to a recurrent network that can learn the decrease along the gradients of the network, and modify the output weights in accordance. We also plan to test our approach on common datasets such as the Pascal VOC Challenge or the Cityscapes Dataset~\cite{cordts2016cityscapes}. As the use of ReLU neurons appeared to pose problems to the combination of weight learning (as in \emph{KGC-weighting} or {\auxnet}) and the Adam optimizer, a next logical step would be to test other activation functions for the DCNN, such as leaky ReLUs.
Lastly, we are planning to investigate further why training on the pickup error was unsuccessful, and aim to modify the function such that it still reflects pickup probabilities, but can also be used as a better scaled version of the distance error for KGC-weighting.


\bibliography{objdetect}

\end{document}